\documentclass[twoside,11pt]{article}

\usepackage[preprint]{jmlr2e}
\usepackage{amsmath}
\usepackage{booktabs}
\usepackage{microtype}
\usepackage{subcaption}
\usepackage{placeins}
\usepackage{ragged2e}
\usepackage{nicefrac}
\usepackage{xcolor}
\usepackage{float}
\usepackage{array}
\usepackage{longtable}
\usepackage{pdflscape}
\graphicspath{{figs_protein/}{figs_molecule/}}

\title{Can Tabular In-Context Learners Generalize to Biomolecular Property Prediction?}

\author{%
    \name Davy Guan$^{1,2}$,
    \name Lu Zhang$^{1}$,
    \name Asiri Wijesinghe$^{1}$,
    \name Allen Zhu$^{1}$,
    \name He Zhao$^{1}$, \\
    \name Helen Power$^{1}$,
    \name F.\ Hafna Ahmed$^{1}$,
    \name Andrew Warden$^{1}$,
    \name Cheng Soon Ong$^{1}$,\\
    \name Daniel M.\ Steinberg$^{1}$ \\
    \addr $^1$Commonwealth Scientific and Industrial Research Organisation (CSIRO), Australia\\
    \addr $^2$Computational Pharmacology and Toxicology Laboratory, Sydney Pharmacy School, The University of Sydney \\
    \email \textnormal{Contact:} dgua4501@uni.sydney.edu.au, dan.steinberg@csiro.au
}

\begin{document}

\maketitle

% Unnumbered first-page note on the venue (arXiv preprint only).
\begingroup
\renewcommand{\thefootnote}{}
\makeatletter
\long\def\@makefntext#1{\noindent #1}% jmlr2e would otherwise print a stray ". " mark
\makeatother
\footnotetext{Accepted at the Asian Conference on Machine Learning (ACML 2026).}
\endgroup

\begin{abstract}
Predicting biomolecular properties from limited labeled data is a central bottleneck in protein engineering and small-molecule design. As strong pretrained encoders now supply rich fixed-length representations, the difficulty has shifted from representation learning to building a data-efficient predictor for the few-shot regime. Tabular foundation models such as TabPFN and TabICL perform this prediction through in-context inference after pretraining on synthetic tables. We test whether they can act as flexible interpolators over informative fixed biomolecular representations. Treating each method as a predictor-representation pair, we evaluate across two domains.
On ProteinGym, TabPFN3-ESMC600M is competitive with leading supervised-substitution entries and obtains the lowest MSE among the leaderboard-context rows considered, while Kermut remains stronger by Spearman. Performance is substantially lower under modulo and contiguous holdouts than under random splits. On the five-endpoint PpEST dataset, tabular models show favorable descriptive performance against matched supervised baselines, but the random split measures within-family interpolation and the small endpoint count precludes strong model-ordering claims.
%For protein fitness regression, ESM-Cambrian (600M) mutation-aware representations improve TabPFN on ProteinGym, but then these larger sequence-level representations do not improve rank correlation over a 300M representation model on a diverse esterase dataset. 
For small-molecule classification with ECFP/RDKit descriptors, no single predictor-representation pairing dominates across TDC ADMET, MoleculeNet, FS-Mol, and DrugOOD, but several pairings are competitive with strong descriptor and graph baselines.
Crucially, on both protein and small-molecule \emph{few-shot} tasks, these predictor-representation pairs offer strong performance.
We conclude that tabular foundation models can be strong biomolecular predictors, but only when coupled with expressive representations.
\end{abstract}


\section{Introduction}
Predicting the functional consequences of biomolecular data from limited labels is a central problem in protein engineering, variant-effect prediction, and small-molecule design. Deep mutational scanning (DMS) benchmarks such as ProteinGym have greatly improved the scale and rigor of evaluation. In reality however,  each new assay typically induces a distinct low-data supervised task in which only a small number of experimentally measured variants are available for a new protein or condition \citep{notin2023proteingym,elnaggar2022prottrans}. Recently, large biological foundation models have made sequence representation substantially stronger: protein language models trained on evolutionary corpora provide rich fixed-length embeddings that can be reused across downstream tasks \citep{elnaggar2022prottrans,lin2023evolutionary,hayes2025simulating}. This shifts part of the modeling bottleneck from representation learning to the design of a data-efficient predictor that can exploit those embeddings in the few-shot regime.

Tabular foundation models offer a compelling approach to this bottleneck. Rather than training a new predictor from scratch for each dataset, they amortize Bayesian-style prediction offline over synthetic tabular tasks and perform inference by conditioning on labeled support examples at test time, without task-specific gradient updates \citep{hollmann2023tabpfn,hollmann2025tabpfn,qu2025tabicl,qu2026tabiclv2}. Recent work has extended this paradigm to larger contexts, broader task families, and more realistic data regimes. Science-domain evaluations in metabolomics, analytical chemistry, and materials informatics suggest that these models can transfer effectively once domain-specific structure has been encoded as feature vectors \citep{wu2026panmetai,granitto2026voc,li2026iclmaterials}.

It remains an open question whether this transfer holds for two domains with distinct representation challenges and output types: protein fitness regression and small-molecule property classification. Protein fitness prediction requires mapping high-dimensional sequence embeddings to continuous fitness values across hundreds of structurally diverse assays with as few as $\mathcal{O}(10)$ labeled examples. Small-molecule property classification poses a different challenge: representations range from fixed fingerprint vectors to learned molecular graph embeddings, and evaluation must simultaneously capture in-distribution accuracy, low-shot sample efficiency, and out-of-distribution generalization across scaffold, assay, and molecular-size shifts.

We study both settings as an evaluation problem using existing pretrained models and fingerprinting techniques. For protein fitness regression, we encode sequences with ESM-Cambrian \citep[referred to as ESMC]{hayes2025simulating} and benchmark TabICL (version 2) and TabPFN (version 3) against supervised baselines across ProteinGym and a newly published diverse esterase family dataset \citep{ahmed2026data}. For small-molecule property classification, we evaluate predictor-representation pairs, pairing tabular models with molecular descriptor views and comparing against graph-based baselines. We benchmark across TDC ADMET \citep{huang2021therapeutics}, MoleculeNet \citep{wu2018moleculenet}, FS-Mol \citep{stanley2021fsmol}, and DrugOOD \citep{ji2023drugood}, with DrugOOD providing an explicit out-of-domain (OOD) generalization test.

Our contribution is primarily empirical and methodological. We provide a benchmarked audit of tabular foundation models in scientific prediction settings that have not yet been evaluated systematically, and we do so under protocols designed to support cautious, track-appropriate claims. We keep feature pipelines fixed where possible, separate few-shot from full-train conclusions, and treat OOD behavior, official ProteinGym holdouts, benchmark coverage, and support-set sensitivity as first-class parts of the evaluation.

\section{Background and Related Work}
The prior-data fitted network (PFN) framework trains a transformer offline on synthetic tabular tasks so that prediction can be performed at inference time by conditioning on labelled in-context examples, without task-specific gradient updates \citep{mullertransformers2022}. TabPFN established this formulation for small classification problems and showed that amortized inference over synthetic Bayesian priors can match or exceed carefully tuned baselines on real benchmarks \citep{hollmann2023tabpfn,hollmann2025tabpfn}. The version 3, 2026 release \citep{grinsztajn2026tabpfn3}, referred to throughout this paper as TabPFN3, extends the framework with improved scalability and regression support. TabICL scales the paradigm through a two-stage architecture: a first stage applies column-then-row attention to compress each sample into a fixed-dimension row embedding, and a second stage applies a transformer across row embeddings for in-context prediction \citep{qu2025tabicl,qu2026tabiclv2}. TabFM is a recent zero-shot tabular foundation model that also treats train and query rows as an in-context prediction problem rather than fitting dataset-specific weights \citep{tabfm2026}. Because it combines row/column attention, row-level compression, and a transformer over compressed row embeddings, we include TabFM on PpEST as a targeted diagnostic of whether newer tabular ICL architectures alone improve biomolecular transfer. We refer to the TabICL v2 implementation used in this study as TabICL for readability.

For protein prediction tasks, ProteinGym standardizes evaluation across 217 DMS assays spanning diverse proteins and fitness phenotypes, enabling systematic comparison across assay types and substitution depths \citep{notin2023proteingym}. The dominant supervised strategy pairs a large pretrained sequence encoder with a lightweight task-specific predictor. Protein language models including ProtTrans \citep{elnaggar2022prottrans}, ESM2 \citep{lin2023evolutionary}, and ESMC \citep{hayes2025simulating} are trained on evolutionary sequence corpora so that their embeddings capture structural and functional constraints learned from natural variation.

Small-molecule property prediction spans ADMET profiling, activity prediction, and out-of-distribution generalization across chemical series. MoleculeNet \citep{wu2018moleculenet}, TDC ADMET \citep{huang2021therapeutics}, FS-Mol \citep{stanley2021fsmol}, and DrugOOD \citep{ji2023drugood} provide complementary regimes covering standard supervised settings, few-shot learning with task-level variation, and systematic distribution shift across scaffolds, assays, and molecular sizes. Fixed descriptors such as ECFP fingerprints \citep{rogers2010extended} and RDKit features expose tabular feature vectors, while graph models such as ChemProp \citep{yang2019analyzing} and CheMeleon-initialized ChemProp \citep{burns2025chemeleon} model molecular structure directly.

\section{Using Tabular ICL Models for Biomolecules}
\label{sec:biomol}
Across both domains we follow a single recipe. A frozen domain encoder maps each biomolecule to a fixed-length feature vector, and that vector becomes one row of an input table for a tabular in-context learner. The predictors, TabICL and TabPFN3, predict the label of a query molecule by conditioning on labeled support rows in context. Protein fitness is treated as regression and reported by MSE and Spearman rank correlation. Small-molecule property prediction is treated as binary classification and reported by ROC-AUC.

\paragraph{Protein representations.}
Protein sequences are encoded with ESMC \citep{hayes2025simulating}, a successor to ESM2 \citep{lin2023evolutionary}. We use ESMC300M sequence embeddings as the default protein feature set. We additionally test ESMC600M in two roles: mutation-aware features for ProteinGym and sequence-level embeddings for a new diverse esterase dataset, PpEST \citep{ahmed2026data}. The ProteinGym ESMC600M rows augment the mutant mean embedding with wildtype-relative mean deltas, absolute deltas, mutation-site token features, simple mutation descriptors, and zero-shot scores. This distinction matters because ProteinGym assays are single-protein mutation tables, whereas PpEST compares diverse homologous sequences across assay conditions. For TabICL on the high-dimensional ProteinGym ESMC600M matrix, we use PCA500 because full-dimensional inference exceeded available memory or wall-time limits.

\paragraph{Molecule features.}
For small molecules, the tabular learners receive either ECFP/Morgan fingerprints \citep{rogers2010extended}, RDKit two-dimensional physicochemical descriptors, or their concatenation. The choice of descriptor is an integral part of the model pair since a tabular learner with ECFP features and the same learner with RDKit descriptors can behave very differently. As structural alternatives to fixed descriptors, graph baselines consume molecular graphs directly through ChemProp \citep{yang2019analyzing} trained from scratch and a foundation-model-initialized variant that fine-tunes ChemProp from the CheMeleon checkpoint \citep{burns2025chemeleon}.

\section{Protein Prediction Experiments}
We evaluate TabICL and TabPFN3 in two protein prediction settings. The first is ProteinGym's 217 DMS tasks, and the second is the PpEST esterase-family dataset introduced in \citet{ahmed2026data}. We test these models in full-training tasks and few-shot settings with 8 to 64 training samples to approximate low-throughput wet-lab measurement regimes.

\paragraph{Datasets.}
\textbf{ProteinGym} \citep{notin2023proteingym} is a large-scale collection of DMS assays curated to benchmark computational models of protein fitness. Each assay quantifies the effect of amino acid substitutions on a biologically relevant property. We use the 217 supervised-substitution assays available in the staged evaluation files and report both random 5-fold results and the official random, modulo, and contiguous holdout schemes. \textbf{PpEST} is a diverse esterase dataset containing 1513 ancestrally related sequences after excluding approximately 600 sequences partially selected using ML methods. It includes catalytic reaction rates for three ester substrates and thermostability measurements at two temperatures \citep{ahmed2026data}.

\paragraph{Baselines.}
We compare TabICL and TabPFN3 with ridge regression, an RBF sampler, HistGradientBoostingRegressor (HGBR) \citep{scikit-learn}, and fine-tuned ESMC with a task-specific MLP regression head. The matched classical baselines receive the same ESMC embeddings. Ridge uses a fixed $\alpha=1$, and the RBF diagnostic uses 512 random Fourier features with fixed $\gamma=1$ followed by the same ridge model; neither underwent a hyperparameter search and neither is treated as a strong tuned baseline. HGBR is the strongest matched classical baseline. The ProteinGym FT ESM row is retained only as a historical aggregate reference because it was not rerun under the current protocol.
%A compatible retrained sequence CNN for PpEST is reported as an appendix check rather than inserted into the main comparison.

\paragraph{Evaluation metrics.}
Protein fitness prediction is evaluated by mean squared error (MSE) and the Spearman rank correlation coefficient. Spearman is the primary ranking metric for comparison with ProteinGym-style variant-effect evaluation, while MSE captures calibrated regression error on the assay-standardized scale. Few-shot figures and tables report the raw mean at each support size rather than a monotone envelope.

For primary comparisons, repeated folds, support draws, or seeds are first averaged within each assay or endpoint. We then compute paired task-level deltas, defined so that positive values favor the first method, including reversed MSE differences so that positive means lower error. We report bootstrap 95\% confidence intervals for the mean delta and two-sided Wilcoxon signed-rank tests. Benjamini-Hochberg correction is applied within each figure or table family. Figure stars denote corrected q-values: $*$ $q<0.05$, $**$ $q<0.01$, and $***$ $q<0.001$. Exact sign tests and complete paired results are reported in the appendix.

\paragraph{Experimental settings.}
In the full-train regime, about 20\% of samples are selected for testing. For ProteinGym, we use five-fold cross-validation and report the mean score averaged across assays. For PpEST, we randomly select 20\% of samples as the test set; this evaluates interpolation within the sampled sequence family, not transfer to distant evolutionary lineages. In the few-shot regime, training sizes are 8, 16, 32, and 64. We run 30 repeated support-set draws for each size using stratified quantile-bin sampling to encourage coverage of the observed fitness range. Because this retrospective protocol uses labels to construct balanced supports, it should not be interpreted as label-agnostic prospective acquisition.

\subsection{ProteinGym Results}
Table~\ref{tab:pg-current-acml} reports model performance on ProteinGym under the random 5-fold cross-validation protocol used for the main reproduced comparison.

\begin{table}[!htbp]
\centering
\small
\begin{tabular}{lcc}
\toprule
\textbf{Model} & \textbf{Spearman} & \textbf{MSE} \\
\midrule
TabPFN3 & \textbf{0.767 $\pm$ 0.172} & \textbf{0.351 $\pm$ 0.251} \\
TabICL & 0.753 $\pm$ 0.178 & 0.376 $\pm$ 0.259 \\
\midrule
Fixed-encoder ESMC300M MLP & 0.722 & 0.405 \\
Training-fold-tuned Ridge & 0.712 & 0.424 \\
HistGradientBoosting & 0.700 $\pm$ 0.189 & 0.439 $\pm$ 0.259 \\
\bottomrule
\end{tabular}
\caption{ProteinGym random 5-fold full-data summary across 217 assays using ESMC300M features. Values are mean $\pm$ assay-level standard deviation where per-assay rows are available. The MLP and tuned Ridge are matched controls fit using training-fold data; the original fixed Ridge and RBF rows are retained only as diagnostics in the appendix.}
\label{tab:pg-current-acml}
\end{table}

TabPFN3 achieves the strongest aggregate result, with mean Spearman 0.767 and mean MSE 0.351 across 217 assays. Against TabICL, the paired mean differences are $\Delta$Spearman $=0.015$ (95\% CI $[0.013,0.019]$, $q=1.3\times10^{-33}$, 206/217 assays favor TabPFN3) and $\Delta$MSE $=0.032$ (95\% CI $[0.024,0.042]$, $q=7.2\times10^{-34}$). Against the fixed-encoder MLP, its paired advantages are 0.045 Spearman and 0.055 MSE after correction; the corresponding advantages over tuned Ridge are 0.055 and 0.074. The two largest assays, HIS7\_YEAST\_Pokusaeva\_2019 and SPG1\_STRSG\_Olson\_2014, required PCA rescue settings. Excluding both leaves the ordering unchanged: TabPFN3/TabICL Spearman is 0.765/0.751 and MSE is 0.353/0.379 across the remaining 215 assays.

\begin{table}[!htbp]
\centering
\begingroup
\footnotesize
\setlength{\tabcolsep}{3.5pt}
\begin{tabular}{lllcc}
\toprule
\textbf{Scheme} & \textbf{Model} & \textbf{Feature set} & \textbf{Spearman} & \textbf{MSE} \\
\midrule
Overall & TabPFN3 & ESMC300M & 0.620 $\pm$ 0.206 & 0.583 $\pm$ 0.461 \\
Overall & TabPFN3 & ESMC600M & \textbf{0.662} $\pm$ 0.195 & 0.537 $\pm$ 0.424 \\
Overall & TabICL & ESMC300M & 0.607 $\pm$ 0.210 & 0.603 $\pm$ 0.475 \\
Overall & TabICL & ESMC600M-PCA500 & 0.625 $\pm$ 0.202 & 0.607 $\pm$ 0.519 \\
\midrule
Random & TabPFN3 & ESMC300M & 0.744 $\pm$ 0.177 & 0.382 $\pm$ 0.257 \\
Random & TabPFN3 & ESMC600M & \textbf{0.782} $\pm$ 0.163 & 0.338 $\pm$ 0.249 \\
Random & TabICL & ESMC300M & 0.729 $\pm$ 0.181 & 0.407 $\pm$ 0.266 \\
Random & TabICL & ESMC600M-PCA500 & 0.773 $\pm$ 0.166 & 0.357 $\pm$ 0.283 \\
Modulo & TabPFN3 & ESMC300M & 0.595 $\pm$ 0.177 & 0.652 $\pm$ 0.544 \\
Modulo & TabPFN3 & ESMC600M & \textbf{0.633} $\pm$ 0.170 & 0.610 $\pm$ 0.498 \\
Modulo & TabICL & ESMC300M & 0.580 $\pm$ 0.182 & 0.669 $\pm$ 0.561 \\
Modulo & TabICL & ESMC600M-PCA500 & 0.584 $\pm$ 0.166 & 0.702 $\pm$ 0.622 \\
Contiguous & TabPFN3 & ESMC300M & 0.522 $\pm$ 0.199 & 0.715 $\pm$ 0.464 \\
Contiguous & TabPFN3 & ESMC600M & \textbf{0.570} $\pm$ 0.187 & 0.664 $\pm$ 0.410 \\
Contiguous & TabICL & ESMC300M & 0.511 $\pm$ 0.205 & 0.732 $\pm$ 0.484 \\
Contiguous & TabICL & ESMC600M-PCA500 & 0.517 $\pm$ 0.179 & 0.762 $\pm$ 0.500 \\
\bottomrule
\end{tabular}
\endgroup
\caption{ProteinGym supervised-substitution performance by official holdout scheme across 217 assays. Values are mean $\pm$ assay-level standard deviation; overall rows summarize all assay-scheme cells. ESMC600M combines mutant mean embeddings, wildtype-relative mean and absolute deltas, mutation-site token features, simple mutation descriptors, and zero-shot scores. TabICL uses PCA500 for computational feasibility, so ESMC600M rows are practical configurations rather than a dimension-matched architecture ablation. Bold marks the best Spearman per scheme; for MSE, lower is better.}
\label{tab:pg-official-folds-acml}
\end{table}

Table~\ref{tab:pg-official-folds-acml} extends the comparison to the official random, modulo, and contiguous ProteinGym fold schemes and adds the ESMC600M representation diagnostic. These values are not directly comparable to Table~\ref{tab:pg-current-acml}: Table~\ref{tab:pg-current-acml} reports our reproduced random 5-fold full-data comparison with ESMC300M features, whereas Table~\ref{tab:pg-official-folds-acml} uses the staged ProteinGym official fold columns and includes representation variants. The harder modulo and contiguous splits reduce performance for all configurations, indicating that random splits primarily measure a more favorable interpolation regime. Across assays and schemes, ESMC600M versus ESMC300M improves TabPFN3 by $\Delta$Spearman $=0.042$ (95\% CI $[0.036,0.047]$, $q=1.9\times10^{-31}$) and $\Delta$MSE $=0.046$ (95\% CI $[0.038,0.055]$, $q=3.7\times10^{-25}$). A fixed-encoder ablation shows that adding zero-shot scores to the mutation-aware ESMC600M features improves MSE overall ($\Delta=0.016$, $q=3.4\times10^{-19}$), but its Spearman effect is split-dependent: small gains under random and modulo splits contrast with a $-0.024$ mean delta under the contiguous split. The remaining ESMC300M-to-ESMC600M difference still conflates encoder and mutation-feature changes and is not interpreted as an encoder-scale ablation. A matched TabPFN3 control further shows that full 9303-dimensional features outperform PCA500 (0.662 versus 0.628 Spearman; 0.537 versus 0.601 MSE), confirming that PCA500 is a feasibility compromise rather than a neutral transformation.

\begin{table}[!htbp]
\centering
\begingroup
\footnotesize
\setlength{\tabcolsep}{3.5pt}
\begin{tabular}{lcc}
\toprule
\textbf{Model} & \textbf{Overall Spearman} & \textbf{Overall MSE} \\
\midrule
Kermut~\citep{groth2024kermut} & \textbf{0.657} & 0.605 \\
\emph{TabPFN3-ESMC600M} & 0.642 & \textbf{0.591} \\
ProteinNPT~\citep{notin2023proteinnpt} & 0.619 & 0.687 \\
\emph{TabPFN3-ESMC300M} & 0.612 & 0.643 \\
MSA Transformer~\citep{rao2021msa} & 0.581 & 0.724 \\
\bottomrule
\end{tabular}
\endgroup
\caption{Overall ProteinGym supervised-substitution context for the top-ranking entries and the two TabPFN3 configurations. Leaderboard-context rows are sourced from ProteinGym benchmark outputs \citep{proteingym_benchmark_page}; values use ProteinGym benchmark aggregation, so they need not equal plain assay means. Higher Spearman and lower MSE are better.}
\label{tab:pg-submission-context-acml}
\end{table}

Table~\ref{tab:pg-submission-context-acml} places the two TabPFN3 configurations in leaderboard context. TabPFN3-ESMC600M improves substantially over TabPFN3-ESMC300M and achieves the lowest overall MSE among these rows, while Kermut remains the strongest overall Spearman reference.

\begin{figure}[!htbp]
\centering
\includegraphics[width=0.84\textwidth]{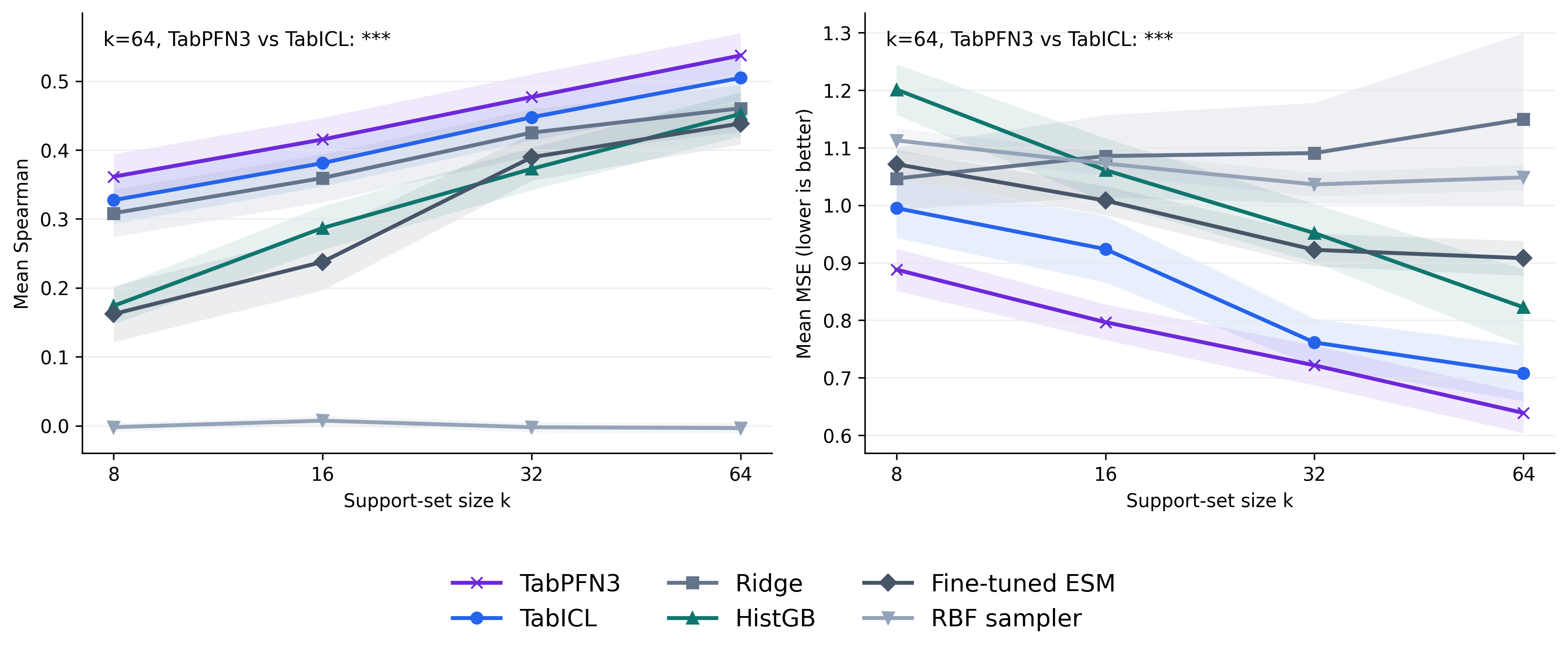}
\caption{ProteinGym few-shot performance as support-set size increases. Curves show raw assay-averaged values and shaded 95\% confidence intervals across assays. The sparse endpoint annotations compare TabPFN3 with TabICL at $k=64$ using the corrected q-value convention defined in Section~4.}
\label{fig:pg-fewshot-acml}
\end{figure}

\begin{table}[!htbp]
\centering
\small
\begin{tabular}{@{}lcccc@{}}
\toprule
Model & Spearman k=8 & MSE k=8 & Spearman k=64 & MSE k=64 \\
\midrule
TabPFN3 & 0.361 $\pm$ 0.247 & 0.888 $\pm$ 0.276 & 0.537 $\pm$ 0.244 & 0.639 $\pm$ 0.262 \\
TabICL & 0.327 $\pm$ 0.255 & 0.995 $\pm$ 0.387 & 0.505 $\pm$ 0.250 & 0.708 $\pm$ 0.362 \\
Ridge & 0.308 $\pm$ 0.259 & 1.046 $\pm$ 0.402 & 0.461 $\pm$ 0.257 & 1.150 $\pm$ 1.122 \\
HistGB & 0.174 $\pm$ 0.195 & 1.201 $\pm$ 0.330 & 0.452 $\pm$ 0.239 & 0.823 $\pm$ 0.498 \\
Fine-tuned ESM & 0.162 $\pm$ 0.306 & 1.071 $\pm$ 0.187 & 0.438 $\pm$ 0.228 & 0.908 $\pm$ 0.226 \\
RBF sampler & -0.002 $\pm$ 0.050 & 1.113 $\pm$ 0.162 & -0.003 $\pm$ 0.054 & 1.048 $\pm$ 0.166 \\
\bottomrule
\end{tabular}

\caption{Raw ProteinGym few-shot summary at the smallest and largest support sizes. Values are mean $\pm$ standard deviation across assays; MSE is computed on the assay-standardized target scale. The $k=64$ aggregate contains 216 assays because one assay has only 63 variants.}
\label{tab:pg-fewshot-tabpfn3-acml}
\end{table}

Figure~\ref{fig:pg-fewshot-acml} and Table~\ref{tab:pg-fewshot-tabpfn3-acml} show raw Spearman and MSE as a function of the number of training samples $k\in\{8,16,32,64\}$. At $k=64$, TabPFN3 exceeds TabICL by mean $\Delta$Spearman $=0.033$ (95\% CI $[0.021,0.044]$, $q=5.6\times10^{-9}$) and $\Delta$MSE $=0.069$ (95\% CI $[0.045,0.099]$, $q=5.1\times10^{-11}$). A label-agnostic five-draw random-support control retains the $k=64$ ordering ($\Delta$Spearman $=0.025$, $\Delta$MSE $=0.043$), while $k=8$ Spearman slightly favors TabICL; full results are in the appendix. The raw curves also expose non-monotone baseline behavior that the previous best-so-far envelope concealed.

\FloatBarrier

\subsection{PpEST Results}
Table~\ref{tab:ppest-acml} reports mean full-training results across the five PpEST endpoints, compares ESMC300M with ESMC600M feature sets, retains the original no-PCA TabICL ESMC300M row, and includes TabFM \citep{tabfm2026} only as a PpEST diagnostic because it was not competitive enough to carry forward into the ProteinGym experiments.

\begin{table}[!htbp]
\centering
\begingroup
\small
\setlength{\tabcolsep}{4pt}
\begin{tabular}{llccc}
\toprule
\textbf{Model} & \textbf{Feature set} & \textbf{PCA} & \textbf{Spearman} & \textbf{MSE} \\
\midrule
TabPFN3 & ESMC300M & none & \textbf{0.661 $\pm$ 0.065} & 0.640 $\pm$ 0.106 \\
TabPFN3 & ESMC600M & none & 0.658 $\pm$ 0.060 & \textbf{0.631 $\pm$ 0.098} \\
TabICL & ESMC300M & none & 0.647 $\pm$ 0.074 & 0.637 $\pm$ 0.097 \\
TabICL & ESMC300M & 500 & 0.643 $\pm$ 0.074 & 0.687 $\pm$ 0.107 \\
TabICL & ESMC600M & 500 & 0.625 $\pm$ 0.079 & 0.678 $\pm$ 0.072 \\
TabFM & ESMC300M & 32 & 0.648 $\pm$ 0.078 & 0.637 $\pm$ 0.099 \\
TabFM & ESMC600M & 32 & 0.644 $\pm$ 0.072 & 0.645 $\pm$ 0.097 \\
TabFM & ESMC300M & 500 & 0.642 $\pm$ 0.086 & 0.639 $\pm$ 0.073 \\
TabFM & ESMC600M & 500 & 0.649 $\pm$ 0.079 & 0.642 $\pm$ 0.081 \\
\midrule
Fine-tuned ESM & ESMC300M & none & 0.588 $\pm$ 0.088 & 0.690 $\pm$ 0.114 \\
Sequence CNN & sequence & none & 0.522 $\pm$ 0.062 & 0.700 $\pm$ 0.139 \\
\bottomrule
\end{tabular}
\endgroup
\caption{Full-training performance averaged over the five PpEST endpoints under the same fixed random 80/20 split. Values are mean $\pm$ endpoint-level standard deviation. The Sequence CNN is a compatible retraining on the five direct endpoints. Because $n=5$, these comparisons are descriptive rather than evidence of a stable model ordering.}
\label{tab:ppest-acml}
\end{table}

On PpEST, ESMC600M sequence-level features reduce the mean TabPFN3 MSE but do not improve mean Spearman over ESMC300M. Endpoint-level tests detect no configuration difference after correction. For example, TabPFN3-ESMC300M versus no-PCA TabICL has $\Delta$Spearman $=0.013$ with a 95\% CI of $[0.003,0.023]$, but the two-sided Wilcoxon q-value is 0.25 at only five endpoints. The no-PCA versus PCA500 TabICL difference is also non-significant ($q=1.0$). We therefore treat the apparent PCA and TabFM differences as observations, not established findings. Fine-tuned ESM and the compatible Sequence CNN provide task-specific full-train context, but even five wins out of five cannot reach $p<0.05$ in a two-sided exact test at this sample size. In an exploratory CD-HIT identity-clustered split, TabPFN3 and TabICL obtain Spearman 0.543/0.532 and MSE 0.879/0.886 over 25 endpoint-fold evaluations; the appendix reports all matched controls.

\begin{figure}[!htbp]
\centering
\includegraphics[width=0.82\textwidth]{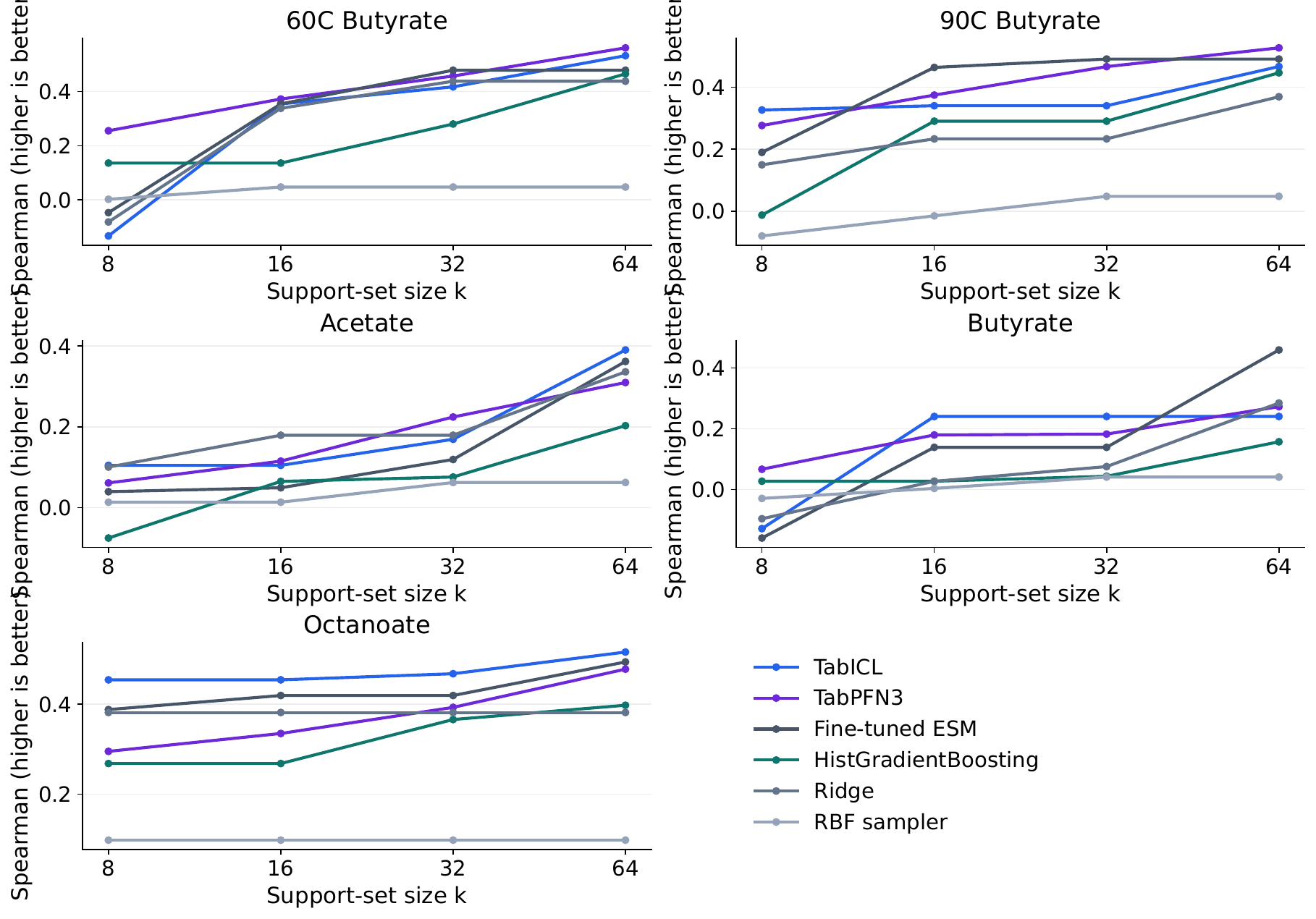}
\caption{Few-shot PpEST performance measured by Spearman correlation. Curves retain the submitted best-so-far visualization, while all quantitative comparisons use raw support-size summaries. Results are exploratory because PpEST contains five endpoints.}
\label{fig:ppest-spearman-acml}
\end{figure}

Figure~\ref{fig:ppest-spearman-acml} shows that TabICL and TabPFN3 exhibit strong rank preservation in the few-shot regime. Across many PpEST endpoints, one of the two tabular foundation models achieves the highest or near-highest Spearman correlation, indicating that they effectively preserve the ordering of protein fitness values even when only a small number of labeled examples are available.

\begin{figure}[!htbp]
\centering
\includegraphics[width=0.82\textwidth]{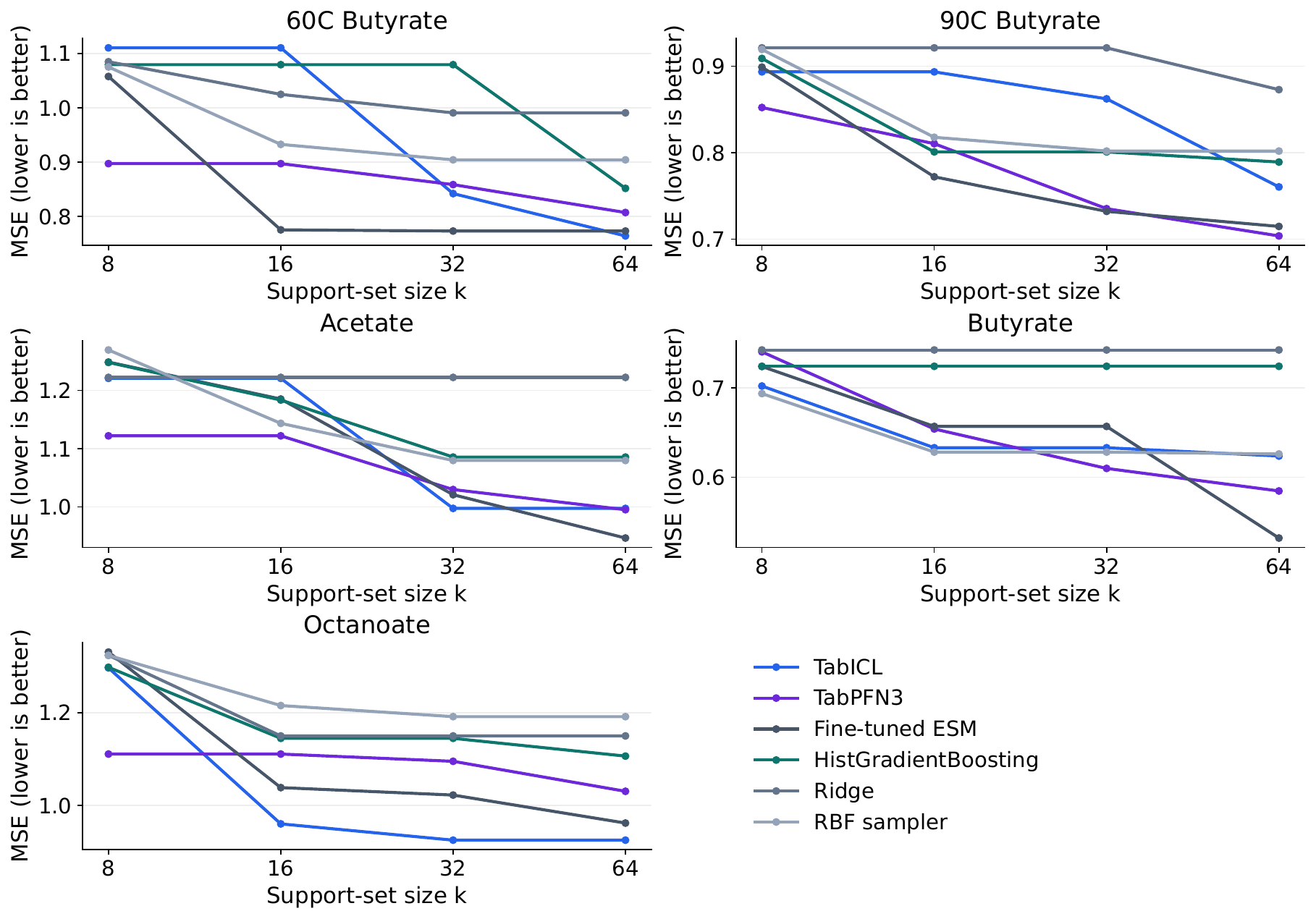}
\caption{Few-shot PpEST performance measured by MSE. Curves retain the submitted best-so-far visualization, while all quantitative comparisons use raw support-size summaries. Results are exploratory because PpEST contains five endpoints.}
\label{fig:ppest-mse-acml}
\end{figure}

Figure~\ref{fig:ppest-mse-acml} shows the corresponding regression-error view. Taken together, PpEST supports a narrower conclusion than ProteinGym: tabular in-context models are effective on ESMC features, but larger sequence embeddings and larger tabular foundation models may not automatically improve rank preservation.

\section{Small-Molecule Prediction Experiments}
We now evaluate whether the behavior of tabular in-context models transfers from protein sequence embeddings to small-molecule classification. The molecular benchmark covers four public benchmark families: TDC ADMET, MoleculeNet, FS-Mol, and DrugOOD-LBAP-core-IC50. We report predictor-representation pairs rather than pure architectures: TabICL, TabPFN3, and XGBoost use fixed molecular descriptors, whereas ChemProp uses molecular graphs and ChemProp+CheMeleon uses graph foundation-model fine-tuning.

\paragraph{Datasets.}
\textbf{TDC ADMET} \citep{huang2021therapeutics} contains 13 curated small-molecule classification endpoints for absorption, distribution, metabolism, excretion, and toxicity prediction. \textbf{MoleculeNet} \citep{wu2018moleculenet} contributes a broad set of molecular classification endpoints, including BBBP, BACE, HIV, ClinTox, SIDER, Tox21, ToxCast, MUV, and PCBA (PubChem BioAssay high-throughput screening endpoints). For multi-label datasets, each label column is treated as a separate binary endpoint. The registry contains 806 MoleculeNet endpoints. Full-train TabPFN3 has 678 completed endpoint outputs per descriptor: 667 defined ROC-AUC values, 11 undefined single-class cases, and no completed outputs for 128 resource-infeasible PCBA endpoints. \textbf{FS-Mol} \citep{stanley2021fsmol} contains 157 few-shot molecular property tasks. \textbf{DrugOOD-LBAP-core-IC50} \citep{ji2023drugood} contains three OOD classification tasks defined by assay, scaffold, and molecular-size shifts.

\begin{table}[!htbp]
\centering
\footnotesize
\begin{tabular}{llcc}
\toprule
\textbf{Benchmark} & \textbf{Role} & \textbf{Tasks/endpoints} & \textbf{Metric} \\
\midrule
TDC ADMET & ADMET classification & 13 & Test ROC-AUC \\
MoleculeNet & Molecular classification & 806 registered & Test ROC-AUC \\
FS-Mol & Few-shot molecular tasks & 157 & Test ROC-AUC \\
DrugOOD & OOD classification & 3 & OOD ROC-AUC \\
\bottomrule
\end{tabular}
\caption{Small-molecule benchmark inventory. MoleculeNet full-train TabPFN3 excludes PCBA endpoints in this snapshot after high-memory rescue attempts proved resource-infeasible.}
\label{tab:mol-inventory-acml}
\end{table}

Table~\ref{tab:mol-inventory-acml} summarizes the molecular benchmark inventory. The key distinction from the protein setting is that the molecular experiments combine several benchmark families with different task construction rules and split semantics, so conclusions are reported at the benchmark-family level rather than collapsed into a single global molecular score.

\paragraph{Baselines and metrics.}
The fixed-feature tabular pairs are TabICL, TabPFN3, and XGBoost \citep{chen2016xgboost}, each paired with ECFP, RDKit, or ECFP+RDKit features. The graph pairs are ChemProp trained from scratch and ChemProp+CheMeleon. The primary metric is ROC-AUC. For TDC ADMET, MoleculeNet, and FS-Mol, we report benchmark test ROC-AUC. For DrugOOD, the main OOD metric is OOD-test ROC-AUC, and we also report the ID/OOD generalization gap, defined as ID-test ROC-AUC minus OOD-test ROC-AUC. Undefined ROC-AUC values can arise when an evaluation split contains a single class; these rows are treated as undefined split/metric cases rather than model failures. For few-shot learning-curve plots, we use the monotone best-so-far envelope,
\begin{align}
    \textrm{AUC}^+_i &= \max \left\{ \textrm{AUC}_i, ~\textrm{AUC}^+_{i-1} \right\}, \quad \textrm{AUC}^+_1 = \textrm{AUC}_1,
\end{align}
where $i$ indexes the few-shot training size, $k \in \{n_1,\ldots,n_i,\ldots,n_I\}$ and $n_i < n_{i+1}$. The envelope is used only for visualization; quantitative comparisons use raw support-size summaries.

\paragraph{Experimental settings.}
In the few-shot regime, support sets are sampled from the training split only, while validation and test splits remain fixed. We use support sizes $k\in\{8,16,32,64,128,256,512\}$ where feasible and five random seeds by default. In full-train experiments, all available training examples are used where feasible. DrugOOD uses train-once/evaluate-many semantics: each fitted model is evaluated on the ID and OOD test splits without split-specific retraining artifacts.

\section{Small-Molecule Results}
\begin{figure}[!htbp]
\centering
\includegraphics[width=0.88\textwidth]{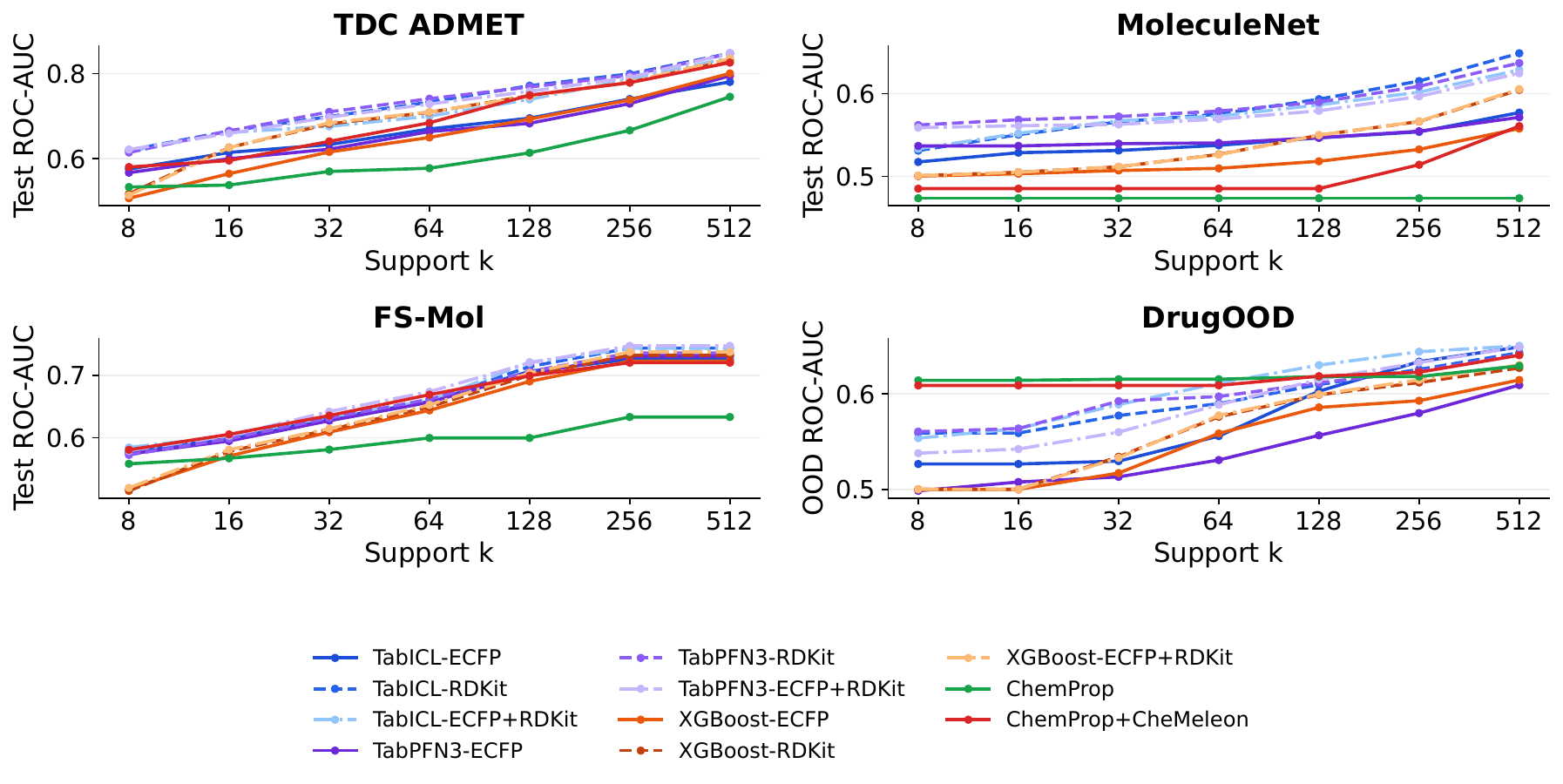}
\caption{Few-shot small-molecule learning curves. The x-axis is the number of labeled support molecules. The y-axis is test ROC-AUC for TDC ADMET, MoleculeNet, and FS-Mol, and OOD-test ROC-AUC for DrugOOD. Curves show monotone best-so-far envelopes of task-averaged performance across support sizes. The envelope is used only for visualization; quantitative comparisons use raw support-size summaries.}
\label{fig:mol-fewshot-acml}
\end{figure}

Figure~\ref{fig:mol-fewshot-acml} shows that fixed-feature tabular foundation models remain competitive in low-label molecular prediction, but the ranking depends strongly on both benchmark family and representation. At the largest support size ($k=512$), TabICL with RDKit descriptors is strongest on MoleculeNet, TabPFN3 with ECFP+RDKit is strongest on FS-Mol, and TabICL RDKit and TabPFN3 ECFP+RDKit are effectively tied on TDC ADMET. On DrugOOD, the best few-shot OOD ROC-AUC values are close among TabICL ECFP+RDKit, TabICL ECFP, and TabPFN3 ECFP+RDKit.

In the full-train regime, TabPFN3 and TabICL are closely matched on the descriptor-based tasks where both are available. Table~\ref{tab:mol-summary-acml} shows that TabPFN3 RDKit slightly exceeds TabICL RDKit on TDC ADMET, while TabPFN3 ECFP+RDKit is strongest on FS-Mol. MoleculeNet full-train TabPFN3 should be interpreted cautiously because PCBA endpoints could not be completed. The corresponding full-train bar plot is given in the appendix.

\begin{table}[!htbp]
\centering
\small
\begin{tabular}{llcc}
\toprule
\textbf{Benchmark} & \textbf{Best full-train pair} & \textbf{ROC-AUC} & \textbf{Coverage} \\
\midrule
TDC ADMET & TabPFN3 RDKit & 0.843 $\pm$ 0.113 & 13/13 \\
MoleculeNet & TabPFN3 / TabICL RDKit & 0.754 / 0.753 & 667/806 each \\
FS-Mol & TabPFN3 ECFP+RDKit & 0.717 $\pm$ 0.135 & 157/157 \\
DrugOOD & ChemProp+CheMeleon & 0.701 $\pm$ 0.018 & 3/3 \\
\bottomrule
\end{tabular}
\caption{Leading full-train pair by benchmark family in the current report snapshot. Values are mean $\pm$ task-level standard deviation where shown. The MoleculeNet comparison uses the exact same 667 endpoints with defined ROC-AUC for TabPFN3 and TabICL; 128 PCBA endpoints are absent from both rows and 11 completed endpoints are undefined.}
\label{tab:mol-summary-acml}
\end{table}

Table~\ref{tab:mol-summary-acml} reinforces the same conclusion: no single predictor-representation pair dominates all molecular benchmark families. On the common 667-endpoint MoleculeNet set, TabPFN3 RDKit and TabICL RDKit differ by only 0.001 mean ROC-AUC (95\% CI $[-0.0027,0.0046]$, $q=0.0575$), so we treat them as practically tied. TabPFN3 ECFP+RDKit is strongest on FS-Mol, while ChemProp+CheMeleon has the highest descriptive mean on the three DrugOOD shifts. DrugOOD has only three shifts and is therefore descriptive: all model families show positive ID-minus-OOD gaps, with magnitudes depending on representation and pretraining regime; the corresponding plot is given in the appendix.

\begin{figure}[!htbp]
\centering
\includegraphics[width=0.72\textwidth]{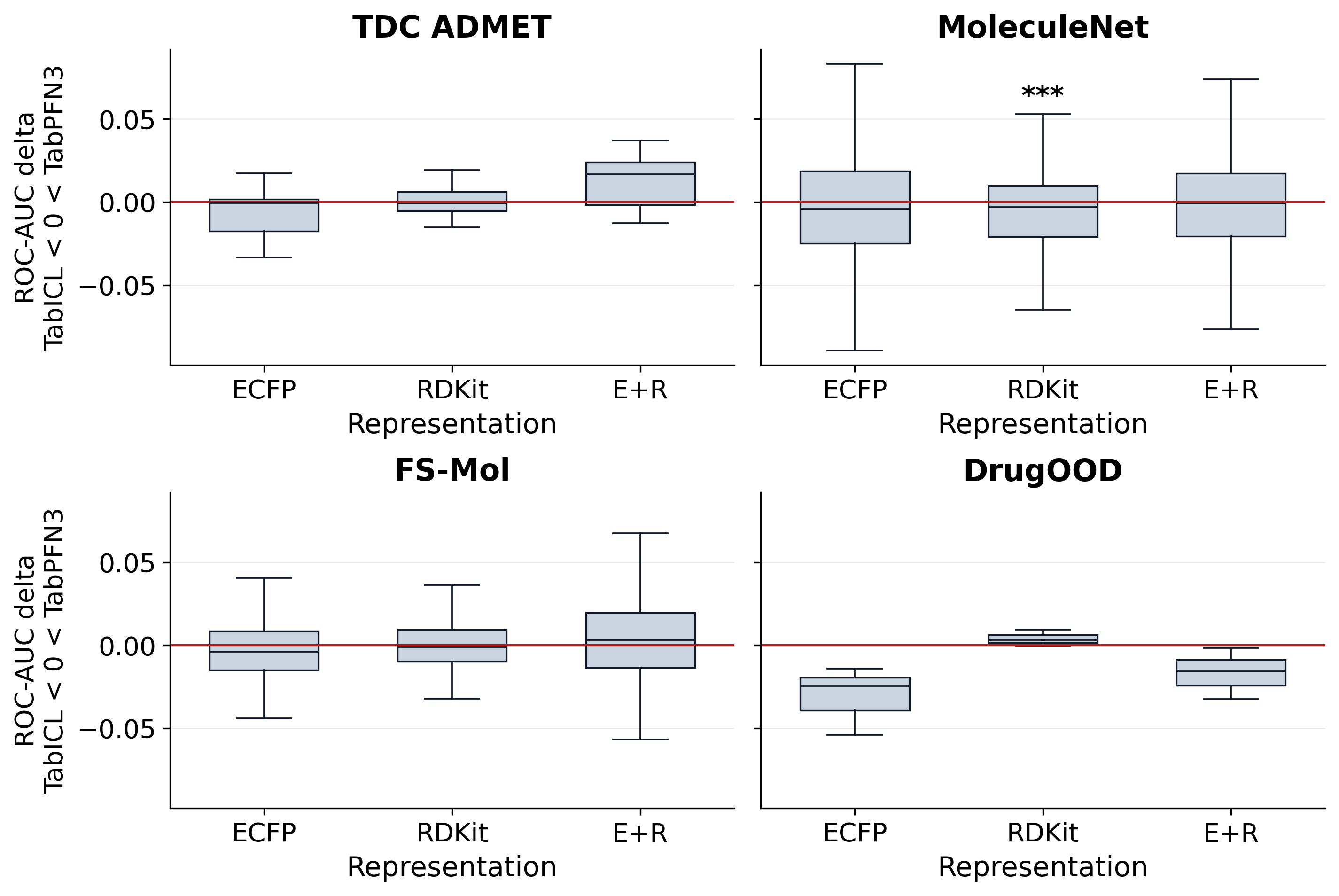}
\caption{Paired few-shot comparison of TabPFN3 and TabICL on molecules. Positive values favor TabPFN3. Each task-level delta first averages matched seeds and support sizes. Stars show two-sided Wilcoxon signed-rank tests after Benjamini-Hochberg correction across the 12 panel-representation cells: $*$ $q<0.05$, $**$ $q<0.01$, $***$ $q<0.001$.}
\label{fig:mol-tabpfn3-delta-acml}
\end{figure}

The paired TabPFN3-TabICL comparison in Figure~\ref{fig:mol-tabpfn3-delta-acml} gives a more controlled view than aggregate ranking alone. Only MoleculeNet with RDKit is significant after correction, favoring TabICL (mean TabPFN3-minus-TabICL delta $=-0.0066$, 95\% CI $[-0.0097,-0.0035]$, $q=1.7\times10^{-5}$). The other 11 cells are not significant after correction. Thus TabPFN3 does not uniformly dominate TabICL, and representation choice and task distribution remain first-order determinants of performance.

\section{Discussion}
The main empirical lesson is that tabular foundation models are useful biomolecular predictors, but not as representation-free black boxes. Their behavior is best understood at the level of a predictor-representation pair. A tabular in-context learner can be strong when the frozen representation already exposes task-relevant neighborhood structure, so that nearby rows in feature space tend to have related labels or rankings. This explains why the same learner can look compelling in one biomolecular setting and merely competitive in another: the predictor is not discovering protein biophysics or chemical structure from raw objects, but exploiting the structure made available by ESMC embeddings, molecular fingerprints, RDKit descriptors, or graph features.

The protein results make this dependence especially clear. On ProteinGym, the complete ESMC600M configuration improves TabPFN3 relative to the earlier ESMC300M configuration, including the lowest aggregate MSE among the leaderboard-context rows considered here. This is a favorable setting for in-context tabular prediction: each assay is a mutation table around a protein background, and the feature representation is explicitly tied to variant effects. The fixed-encoder ablation shows that zero-shot scores improve MSE but have split-dependent effects on Spearman, including a reduction under contiguous holdout. The remaining representation contrast jointly changes encoder scale and mutation features, so it cannot identify a single causal source of improvement. The harder modulo and contiguous ProteinGym splits remain much less forgiving than random folds, which shows that random-split performance is primarily evidence of local interpolation rather than a complete estimate of variant-effect generalization.

PpEST provides a useful counterpoint rather than a contradiction. There, ESMC600M sequence-level features improve TabPFN3 MSE slightly but do not improve the primary Spearman result over ESMC300M. The no-PCA TabICL row also has a slightly higher mean Spearman than PCA500, but neither difference is significant across five endpoints. These descriptive observations suggest hypotheses rather than a model ordering: larger encoders and feature matrices are not guaranteed to help diverse homologous-sequence regression. Moreover, the random split measures within-family interpolation, and the exploratory identity-clustered result should not be interpreted as definitive evidence of distant evolutionary transfer.

The molecular experiments show the same principle in a different domain. TabICL and TabPFN3 are competitive with XGBoost and graph baselines in several few-shot and full-train settings, but no learner dominates across TDC ADMET, MoleculeNet, FS-Mol, and DrugOOD. Descriptor choice changes the ranking within the same model family, and graph-based ChemProp+CheMeleon remains strongest in the DrugOOD full-train summary. This supports a cautious practical recommendation: for small-molecule prediction, tabular foundation models should be benchmarked as descriptor-conditioned predictors, not as replacements for molecular representation design.

The TabFM diagnostic strengthens this interpretation. TabFM is a newer tabular foundation model, yet in the PpEST experiments it does not provide a clean improvement over TabPFN3 under the tested ESMC feature settings. This negative result is useful because it rules out a simplistic ``newer tabular foundation model is better'' story. The decisive factor is not just model scale or recency, but whether the model, representation, and split regime form a compatible prediction problem.

Overall, these results support a more modest but more durable claim than a leaderboard-style model ranking. Tabular in-context learning can be a practical interface between pretrained biological or chemical representations and low-label supervised prediction. Its value is strongest when the representation already organizes examples in a way that makes local interpolation meaningful, and weakest when the feature view, task structure, or distribution shift breaks that assumption. Future benchmarks for in-context learning should therefore report support-set sensitivity, representation choice, and split regime alongside aggregate performance.

\paragraph{Limitations.}
Several limitations should remain explicit. First, the ESMC600M feature update is not a controlled encoder-scale comparison: ProteinGym changes encoder scale and mutation-aware features, whereas PpEST uses plain sequence-level embeddings. Second, TabICL uses PCA500 on ESMC600M because full-dimensional inference was resource-infeasible, so its ESMC600M comparison with TabPFN3 is not dimension matched; the ESMC300M comparison is matched. Third, two of 217 random-fold ProteinGym assays required PCA rescue, although excluding both does not alter the aggregate ordering, and harder modulo and contiguous holdouts reduce all models' performance. Fourth, the random PpEST split does not test distant lineage transfer, and five endpoints provide little inferential power. Fifth, label-aware quantile-bin supports are a retrospective coverage protocol rather than prospective acquisition. Sixth, MoleculeNet full-train TabPFN3 omits 128 PCBA endpoints and has 11 undefined ROC-AUC endpoints among 678 completed outputs; comparisons use common defined endpoints only. Finally, the artifacts do not contain uniform peak-memory telemetry across all method-assay pairs, preventing a complete hardware scaling analysis.

\paragraph{Future directions.}
The most immediate next step is to turn the predictor-representation framing into a diagnostic tool. Rather than asking only which model wins after a full benchmark run, future work should estimate whether a new assay is likely to benefit from TabPFN3, TabICL, a graph model, or a simpler supervised baseline. Useful predictors may include label smoothness in representation, nearest-neighbor predictability, support-set coverage, assay size, feature dimensionality, and the gap between random and structured holdout splits.

A second direction is support-set, representation, and systems design. Active support selection, diversity-aware support construction, and representation-specific diagnostics could make in-context prediction more useful for prospective wet-lab design. For proteins, mutation-aware, sequence-level, structural, and MSA-derived features should be compared under matched supervised protocols. Large DMS assays and PCBA-like molecular screens also expose scalability limits; future systems should support streaming or sharded inference, predictable memory use, and auditable records of feature construction, split IDs, support-set IDs, rescue settings, and complete predictions.

\section{Conclusion}
These results show that tabular foundation models are viable biomolecular predictors when used as part of predictor-representation pairs, but they should not be evaluated as representation-free black boxes. ESMC plus TabPFN3 or TabICL is effective for protein regression, while descriptor-specific TabPFN3 and TabICL pairings are competitive with strong molecular baselines in several settings. Their strongest few-shot results occur under retrospective support-selection protocols and informative representations; structured holdouts, incomplete endpoint coverage, and scalability constraints remain essential qualifications.
%The ESMC600M and TabFM diagnostics reinforce the same point: stronger components help only when they fit the task representation and split regime. 
%This framing makes the positive result more credible and the limitations easier to audit.

\FloatBarrier
\begingroup
\footnotesize
\setlength{\bibsep}{0pt}
\bibliography{ref}
\endgroup

\clearpage
\appendix
% Appendix: adapted from the ACML 2026 camera-ready supplementary materials.
\begingroup
\sloppy
These appendices provide additional protocol details, diagnostic tables, and figures that support reproducibility and interpretation. Exploratory or incomplete diagnostics are clearly marked and are not used as headline evidence.

\section{Feature And Evaluation Details}

Table~\ref{tab:supp-protocol-summary} summarizes the main experimental settings that are easiest to confuse across protein and molecular tasks.

\begin{table}[H]
\centering
\small
\setlength{\tabcolsep}{4pt}
\begin{tabular}{>{\raggedright\arraybackslash}p{0.17\textwidth}>{\raggedright\arraybackslash}p{0.24\textwidth}>{\raggedright\arraybackslash}p{0.25\textwidth}>{\raggedright\arraybackslash}p{0.24\textwidth}}
\toprule
\textbf{Dataset family} & \textbf{Representation} & \textbf{Evaluation splits} & \textbf{Primary metrics} \\
\midrule
ProteinGym & ESMC300M sequence-level features; ESMC600M mutation-aware features for the representation diagnostic & Random 5-fold and official random, modulo, contiguous holdouts & Spearman and MSE \\
PpEST & ESMC300M and ESMC600M sequence-level embeddings & Fixed 80/20 split and few-shot supports & Spearman and MSE \\
TDC ADMET & ECFP, RDKit, ECFP+RDKit descriptors; molecular graphs for graph baselines & Benchmark train/test split & ROC-AUC \\
MoleculeNet & ECFP, RDKit, ECFP+RDKit descriptors; molecular graphs for graph baselines & Benchmark train/test split & ROC-AUC \\
FS-Mol & ECFP, RDKit, ECFP+RDKit descriptors; molecular graphs for graph baselines & Few-shot benchmark tasks & ROC-AUC \\
DrugOOD & ECFP, RDKit, ECFP+RDKit descriptors; molecular graphs for graph baselines & ID and OOD test splits & OOD ROC-AUC and ID/OOD gap \\
\bottomrule
\end{tabular}
\caption{Protocol summary across benchmark families.}
\label{tab:supp-protocol-summary}
\end{table}

The ESMC600M feature update is not a single interchangeable representation across all protein tasks. ProteinGym uses mutation-aware features: mutant-level ESMC600M features augmented with wildtype-relative deltas, absolute deltas, mutation-site features, simple mutation descriptors, and zero-shot scores. PpEST uses sequence-level ESMC600M embeddings because it compares diverse homologous sequences across assay conditions rather than mutations around a single reference sequence.

For TabICL on high-dimensional ProteinGym ESMC600M features, PCA500 is used as a computational feasibility setting. This suffix denotes compression of the feature matrix before fitting TabICL, not a different biological representation. For molecular benchmarks, PCBA denotes PubChem BioAssay high-throughput screening endpoints within MoleculeNet.

\section{Paired Statistical Testing}

Repeated folds, support draws, seeds, and split schemes are first averaged within each assay, task, or endpoint. Paired deltas are then computed across matched tasks, with positive values defined to favor the first method. Thus Spearman and ROC-AUC use score$_A-$score$_B$, while MSE uses MSE$_B-$MSE$_A$. We report the mean and median paired delta, a deterministic 10,000-sample bootstrap 95\% confidence interval for the mean, a two-sided Wilcoxon signed-rank test, and an exact two-sided sign test. Benjamini-Hochberg correction is applied within each figure or table family. PpEST ($n=5$) and DrugOOD ($n=3$) are descriptive only.

\begin{landscape}
\scriptsize
\setlength{\tabcolsep}{1.5pt}
\renewcommand{\arraystretch}{0.94}
\begin{longtable}{>{\raggedright\arraybackslash}p{1.3cm}>{\raggedright\arraybackslash}p{1.8cm}>{\raggedright\arraybackslash}p{2.5cm}>{\centering\arraybackslash}p{0.7cm}>{\raggedright\arraybackslash}p{3.6cm}>{\centering\arraybackslash}p{1.2cm}>{\centering\arraybackslash}p{0.5cm}>{\centering\arraybackslash}p{0.9cm}>{\centering\arraybackslash}p{1.0cm}>{\centering\arraybackslash}p{1.8cm}>{\centering\arraybackslash}p{1.2cm}>{\centering\arraybackslash}p{1.1cm}>{\centering\arraybackslash}p{1.1cm}>{\centering\arraybackslash}p{0.6cm}>{\centering\arraybackslash}p{0.6cm}}
\caption{Primary paired statistical tests. Spearman correlation is denoted by $\rho$. Deltas are task-level paired effects with positive values favoring the first method; for MSE and ID/OOD gap, positive means a lower value for the first method. q-values use Benjamini-Hochberg correction within each figure/table family. Notes: A, assay means; B, official splits (schemes averaged within assay for ``overall''); C, fixed ESMC600M encoder with only zero-shot features added; D, strongest supervised baseline selected within the same metric and support size; E, seeds and support sizes averaged within endpoint; F, molecular few-shot $k=512$ with seeds averaged within endpoint; G, full-train molecular comparison on common defined endpoints; H, descriptive DrugOOD comparison ($n=3$). The pair count $n$ gives the overlapping defined coverage.}\label{tab:supp-primary-stats}\\
\toprule
\textbf{Domain} & \textbf{Bench.} & \textbf{Setting} & \textbf{Met.} & \textbf{Comparison} & \textbf{Unit} & \textbf{n} & \textbf{Mean $\Delta$} & \textbf{Med. $\Delta$} & \textbf{95\% CI} & \textbf{Wilcox. p} & \textbf{Sign p} & \textbf{BH q} & \textbf{Sig.} & \textbf{Key} \\
\midrule
\endfirsthead
\toprule
\textbf{Domain} & \textbf{Bench.} & \textbf{Setting} & \textbf{Met.} & \textbf{Comparison} & \textbf{Unit} & \textbf{n} & \textbf{Mean $\Delta$} & \textbf{Med. $\Delta$} & \textbf{95\% CI} & \textbf{Wilcox. p} & \textbf{Sign p} & \textbf{BH q} & \textbf{Sig.} & \textbf{Key} \\
\midrule
\endhead
Protein & ProteinGym & random 5-fold; ESMC300M & $\rho$ & TabPFN3 vs TabICL & assay & 217 & 0.015 & 0.010 & [0.013, 0.019] & 1.32e-33 & 9.75e-48 & 1.32e-33 & *** & A \\
Protein & ProteinGym & random 5-fold; ESMC300M & $\rho$ & TabPFN3 vs HistGB & assay & 217 & 0.065 & 0.052 & [0.059, 0.071] & 2.35e-37 & 9.50e-66 & 5.71e-37 & *** & A \\
Protein & ProteinGym & random 5-fold; ESMC300M & $\rho$ & TabICL vs HistGB & assay & 217 & 0.050 & 0.040 & [0.045, 0.054] & 2.85e-37 & 8.69e-58 & 5.71e-37 & *** & A \\
Protein & ProteinGym & random 5-fold; ESMC300M & MSE & TabPFN3 vs TabICL & assay & 217 & 0.032 & 0.017 & [0.024, 0.042] & 5.97e-34 & 9.75e-48 & 7.17e-34 & *** & A \\
Protein & ProteinGym & random 5-fold; ESMC300M & MSE & TabPFN3 vs HistGB & assay & 217 & 0.090 & 0.078 & [0.082, 0.099] & 2.38e-37 & 2.07e-63 & 5.71e-37 & *** & A \\
Protein & ProteinGym & random 5-fold; ESMC300M & MSE & TabICL vs HistGB & assay & 217 & 0.058 & 0.055 & [0.052, 0.064] & 1.44e-34 & 3.72e-56 & 2.17e-34 & *** & A \\
Protein & ProteinGym & official; overall & $\rho$ & T3-600M vs T3-300M & assay & 217 & 0.042 & 0.039 & [0.036, 0.047] & 2.39e-32 & 9.43e-38 & 1.91e-31 & *** & B \\
Protein & ProteinGym & official; random & $\rho$ & T3-600M vs T3-300M & assay & 217 & 0.038 & 0.024 & [0.033, 0.044] & 9.68e-35 & 9.75e-48 & 2.32e-33 & *** & B \\
Protein & ProteinGym & official; modulo & $\rho$ & T3-600M vs T3-300M & assay & 217 & 0.039 & 0.037 & [0.033, 0.045] & 1.95e-27 & 6.45e-29 & 3.61e-27 & *** & B \\
Protein & ProteinGym & official; contiguous & $\rho$ & T3-600M vs T3-300M & assay & 217 & 0.048 & 0.045 & [0.041, 0.056] & 4.32e-28 & 1.02e-29 & 1.04e-27 & *** & B \\
Protein & ProteinGym & official; overall & $\rho$ & T3-600M vs TICL-600M-PCA500 & assay & 217 & 0.037 & 0.028 & [0.031, 0.043] & 6.03e-28 & 9.46e-39 & 1.32e-27 & *** & B \\
Protein & ProteinGym & official; random & $\rho$ & T3-600M vs TICL-600M-PCA500 & assay & 217 & 0.010 & 0.007 & [0.006, 0.014] & 2.00e-15 & 1.88e-16 & 2.28e-15 & *** & B \\
Protein & ProteinGym & official; modulo & $\rho$ & T3-600M vs TICL-600M-PCA500 & assay & 217 & 0.049 & 0.035 & [0.040, 0.058] & 1.06e-25 & 6.45e-29 & 1.70e-25 & *** & B \\
Protein & ProteinGym & official; contiguous & $\rho$ & T3-600M vs TICL-600M-PCA500 & assay & 217 & 0.053 & 0.042 & [0.045, 0.061] & 2.87e-28 & 7.98e-36 & 7.66e-28 & *** & B \\
Protein & ProteinGym & official; overall & $\rho$ & T3-300M vs TICL-300M & assay & 217 & 0.013 & 0.011 & [0.011, 0.016] & 4.01e-27 & 1.56e-30 & 6.87e-27 & *** & B \\
Protein & ProteinGym & official; random & $\rho$ & T3-300M vs TICL-300M & assay & 217 & 0.015 & 0.011 & [0.013, 0.018] & 9.19e-32 & 2.47e-50 & 5.52e-31 & *** & B \\
Protein & ProteinGym & official; modulo & $\rho$ & T3-300M vs TICL-300M & assay & 217 & 0.015 & 0.013 & [0.011, 0.018] & 1.21e-21 & 7.03e-20 & 1.71e-21 & *** & B \\
Protein & ProteinGym & official; contiguous & $\rho$ & T3-300M vs TICL-300M & assay & 217 & 0.011 & 0.010 & [0.007, 0.014] & 1.52e-14 & 1.27e-11 & 1.66e-14 & *** & B \\
Protein & ProteinGym & official; overall & MSE & T3-600M vs T3-300M & assay & 217 & 0.046 & 0.041 & [0.038, 0.055] & 2.44e-25 & 4.47e-23 & 3.66e-25 & *** & B \\
Protein & ProteinGym & official; random & MSE & T3-600M vs T3-300M & assay & 217 & 0.045 & 0.032 & [0.038, 0.052] & 1.42e-31 & 6.81e-35 & 6.80e-31 & *** & B \\
Protein & ProteinGym & official; modulo & MSE & T3-600M vs T3-300M & assay & 217 & 0.042 & 0.032 & [0.030, 0.056] & 3.88e-18 & 1.09e-18 & 4.66e-18 & *** & B \\
Protein & ProteinGym & official; contiguous & MSE & T3-600M vs T3-300M & assay & 217 & 0.051 & 0.042 & [0.040, 0.064] & 1.79e-20 & 7.03e-20 & 2.26e-20 & *** & B \\
Protein & ProteinGym & official; overall & MSE & T3-600M vs TICL-600M-PCA500 & assay & 217 & 0.070 & 0.042 & [0.056, 0.086] & 2.00e-28 & 5.54e-34 & 6.87e-28 & *** & B \\
Protein & ProteinGym & official; random & MSE & T3-600M vs TICL-600M-PCA500 & assay & 217 & 0.019 & 0.010 & [0.011, 0.029] & 2.96e-11 & 6.84e-15 & 2.96e-11 & *** & B \\
Protein & ProteinGym & official; modulo & MSE & T3-600M vs TICL-600M-PCA500 & assay & 217 & 0.092 & 0.050 & [0.071, 0.117] & 2.40e-28 & 6.45e-29 & 7.19e-28 & *** & B \\
Protein & ProteinGym & official; contiguous & MSE & T3-600M vs TICL-600M-PCA500 & assay & 217 & 0.098 & 0.065 & [0.080, 0.120] & 1.91e-27 & 7.07e-26 & 3.61e-27 & *** & B \\
Protein & ProteinGym & official; overall & MSE & T3-300M vs TICL-300M & assay & 217 & 0.019 & 0.018 & [0.015, 0.024] & 3.63e-30 & 8.90e-37 & 1.45e-29 & *** & B \\
Protein & ProteinGym & official; random & MSE & T3-300M vs TICL-300M & assay & 217 & 0.024 & 0.018 & [0.020, 0.028] & 1.76e-33 & 1.32e-54 & 2.11e-32 & *** & B \\
Protein & ProteinGym & official; modulo & MSE & T3-300M vs TICL-300M & assay & 217 & 0.017 & 0.016 & [0.009, 0.024] & 9.94e-21 & 4.11e-18 & 1.33e-20 & *** & B \\
Protein & ProteinGym & official; contiguous & MSE & T3-300M vs TICL-300M & assay & 217 & 0.017 & 0.013 & [0.012, 0.023] & 2.05e-13 & 1.98e-13 & 2.14e-13 & *** & B \\
Protein & ProteinGym & 600M ZS ablation; overall & $\rho$ & mutation-aware + ZS vs mutation-aware & assay & 217 & -0.006 & -0.001 & [-0.010, -0.002] & 0.0400 & 0.4154 & 0.0400 & * & C \\
Protein & ProteinGym & 600M ZS ablation; random & $\rho$ & mutation-aware + ZS vs mutation-aware & assay & 217 & 0.003 & 0.002 & [0.001, 0.005] & 2.08e-07 & 1.64e-06 & 3.33e-07 & *** & C \\
Protein & ProteinGym & 600M ZS ablation; modulo & $\rho$ & mutation-aware + ZS vs mutation-aware & assay & 217 & 0.003 & 0.004 & [-8.75e-04, 0.007] & 2.22e-04 & 8.12e-07 & 2.53e-04 & *** & C \\
Protein & ProteinGym & 600M ZS ablation; contiguous & $\rho$ & mutation-aware + ZS vs mutation-aware & assay & 217 & -0.024 & -0.008 & [-0.035, -0.014] & 1.56e-04 & 0.0296 & 2.08e-04 & *** & C \\
Protein & ProteinGym & 600M ZS ablation; overall & MSE & mutation-aware + ZS vs mutation-aware & assay & 217 & 0.016 & 0.006 & [8.66e-04, 0.038] & 8.41e-20 & 1.88e-16 & 3.36e-19 & *** & C \\
Protein & ProteinGym & 600M ZS ablation; random & MSE & mutation-aware + ZS vs mutation-aware & assay & 217 & 0.008 & 0.003 & [0.004, 0.013] & 1.96e-21 & 2.81e-19 & 1.57e-20 & *** & C \\
Protein & ProteinGym & 600M ZS ablation; modulo & MSE & mutation-aware + ZS vs mutation-aware & assay & 217 & 0.030 & 0.007 & [-0.003, 0.084] & 1.92e-17 & 6.40e-16 & 5.11e-17 & *** & C \\
Protein & ProteinGym & 600M ZS ablation; contiguous & MSE & mutation-aware + ZS vs mutation-aware & assay & 217 & 0.011 & 0.008 & [-0.002, 0.022] & 5.99e-12 & 4.64e-12 & 1.20e-11 & *** & C \\
Protein & ProteinGym & few-shot k=8 & $\rho$ & TabPFN3 vs TabICL & assay & 217 & 0.034 & 0.015 & [0.009, 0.059] & 0.0182 & 0.1744 & 0.0182 & * & D \\
Protein & ProteinGym & few-shot k=8 & $\rho$ & TabPFN3 vs Ridge & assay & 217 & 0.053 & 0.036 & [0.026, 0.080] & 2.24e-04 & 0.0414 & 2.68e-04 & *** & D \\
Protein & ProteinGym & few-shot k=8 & $\rho$ & TabICL vs Ridge & assay & 217 & 0.019 & 0.010 & [0.009, 0.029] & 5.77e-04 & 0.0097 & 6.29e-04 & *** & D \\
Protein & ProteinGym & few-shot k=8 & MSE & TabPFN3 vs TabICL & assay & 217 & 0.107 & 0.071 & [0.063, 0.152] & 9.01e-07 & 7.50e-05 & 1.20e-06 & *** & D \\
Protein & ProteinGym & few-shot k=8 & MSE & TabPFN3 vs Ridge & assay & 217 & 0.158 & 0.111 & [0.114, 0.203] & 6.92e-12 & 8.12e-07 & 1.66e-11 & *** & D \\
Protein & ProteinGym & few-shot k=8 & MSE & TabICL vs Ridge & assay & 217 & 0.051 & 0.052 & [0.033, 0.071] & 7.60e-09 & 3.94e-08 & 1.14e-08 & *** & D \\
Protein & ProteinGym & few-shot k=64 & $\rho$ & TabPFN3 vs TabICL & assay & 216 & 0.033 & 0.021 & [0.021, 0.044] & 3.28e-09 & 5.45e-08 & 5.63e-09 & *** & D \\
Protein & ProteinGym & few-shot k=64 & $\rho$ & TabPFN3 vs Ridge & assay & 216 & 0.077 & 0.068 & [0.064, 0.089] & 4.94e-24 & 3.40e-22 & 1.97e-23 & *** & D \\
Protein & ProteinGym & few-shot k=64 & $\rho$ & TabICL vs Ridge & assay & 216 & 0.044 & 0.041 & [0.034, 0.054] & 1.98e-15 & 1.04e-14 & 5.94e-15 & *** & D \\
Protein & ProteinGym & few-shot k=64 & MSE & TabPFN3 vs TabICL & assay & 216 & 0.069 & 0.039 & [0.045, 0.099] & 2.57e-11 & 2.54e-07 & 5.14e-11 & *** & D \\
Protein & ProteinGym & few-shot k=64 & MSE & TabPFN3 vs HistGB & assay & 216 & 0.184 & 0.135 & [0.142, 0.242] & 1.68e-33 & 9.91e-43 & 2.02e-32 & *** & D \\
Protein & ProteinGym & few-shot k=64 & MSE & TabICL vs HistGB & assay & 216 & 0.115 & 0.090 & [0.090, 0.147] & 4.53e-30 & 1.11e-28 & 2.72e-29 & *** & D \\
Small mol. & TDC ADMET & few-shot k=512 & AUC & TabICL RDKit vs XGBoost ECFP+RDKit & endpoint & 6 & 0.013 & 0.013 & [0.007, 0.017] & 0.0312 & 0.0312 & 0.0859 & n.s. & F \\
Small mol. & TDC ADMET & few-shot k=512 & AUC & TabPFN3 ECFP vs TabICL ECFP & endpoint & 6 & 0.014 & 0.011 & [0.003, 0.027] & 0.1562 & 0.6875 & 0.2778 & n.s. & F \\
Small mol. & TDC ADMET & few-shot k=512 & AUC & TabPFN3 RDKit vs TabICL RDKit & endpoint & 6 & -7.19e-04 & -0.001 & [-0.004, 0.003] & 0.8438 & 0.6875 & 0.9000 & n.s. & F \\
Small mol. & TDC ADMET & few-shot k=512 & AUC & TabPFN3 ECFP+RDKit vs TabICL ECFP+RDKit & endpoint & 6 & 0.011 & 0.013 & [0.005, 0.015] & 0.0625 & 0.2188 & 0.1429 & n.s. & F \\
Small mol. & MoleculeNet & few-shot k=512 & AUC & TabICL RDKit vs XGBoost ECFP+RDKit & endpoint & 653 & 0.043 & 0.031 & [0.037, 0.049] & 5.01e-57 & 5.92e-65 & 8.02e-56 & *** & F \\
Small mol. & MoleculeNet & few-shot k=512 & AUC & TabPFN3 ECFP vs TabICL ECFP & endpoint & 653 & -0.004 & -0.010 & [-0.012, 0.005] & 0.0174 & 7.52e-04 & 0.0696 & n.s. & F \\
Small mol. & MoleculeNet & few-shot k=512 & AUC & TabPFN3 RDKit vs TabICL RDKit & endpoint & 653 & -0.010 & -0.007 & [-0.013, -0.006] & 3.72e-13 & 1.87e-10 & 2.98e-12 & *** & F \\
Small mol. & MoleculeNet & few-shot k=512 & AUC & TabPFN3 ECFP+RDKit vs TabICL ECFP+RDKit & endpoint & 652 & 7.17e-04 & 0.006 & [-0.006, 0.007] & 0.0090 & 0.0015 & 0.0477 & * & F \\
Small mol. & FS-Mol & few-shot k=512 & AUC & TabICL RDKit vs XGBoost ECFP+RDKit & endpoint & 11 & 0.006 & 0.015 & [-0.003, 0.015] & 0.1230 & 1.0000 & 0.2461 & n.s. & F \\
Small mol. & FS-Mol & few-shot k=512 & AUC & TabPFN3 ECFP vs TabICL ECFP & endpoint & 11 & 0.006 & 0.006 & [-0.008, 0.018] & 0.5195 & 1.0000 & 0.5938 & n.s. & F \\
Small mol. & FS-Mol & few-shot k=512 & AUC & TabPFN3 RDKit vs TabICL RDKit & endpoint & 11 & -0.008 & -0.007 & [-0.014, -8.27e-04] & 0.0322 & 0.0117 & 0.0859 & n.s. & F \\
Small mol. & FS-Mol & few-shot k=512 & AUC & TabPFN3 ECFP+RDKit vs TabICL ECFP+RDKit & endpoint & 11 & 0.007 & 0.006 & [-0.002, 0.017] & 0.1748 & 0.0654 & 0.2797 & n.s. & F \\
Small mol. & DrugOOD & few-shot k=512 & AUC & TabICL ECFP+RDKit vs ChemProp+CheMeleon & endpoint & 3 & 0.010 & 0.015 & [-0.003, 0.017] & 0.5000 & 1.0000 & 0.5938 & n.s. & F \\
Small mol. & DrugOOD & few-shot k=512 & AUC & TabPFN3 ECFP vs TabICL ECFP & endpoint & 3 & -0.040 & -0.034 & [-0.065, -0.022] & 0.2500 & 0.2500 & 0.3636 & n.s. & F \\
Small mol. & DrugOOD & few-shot k=512 & AUC & TabPFN3 RDKit vs TabICL RDKit & endpoint & 3 & -0.002 & -9.44e-04 & [-0.005, 2.00e-04] & 0.5000 & 1.0000 & 0.5938 & n.s. & F \\
Small mol. & DrugOOD & few-shot k=512 & AUC & TabPFN3 ECFP+RDKit vs TabICL ECFP+RDKit & endpoint & 3 & -0.001 & -0.005 & [-0.005, 0.005] & 1.0000 & 1.0000 & 1.0000 & n.s. & F \\
Small mol. & TDC ADMET & full-train common endpoints & AUC & TabPFN3 RDKit vs TabICL RDKit & endpoint & 13 & 0.002 & 0.001 & [-0.002, 0.006] & 0.3396 & 0.5811 & 0.3396 & n.s. & G \\
Small mol. & TDC ADMET & full-train common endpoints & AUC & TabPFN3 RDKit vs ChemProp+CheMeleon & endpoint & 13 & 0.012 & 0.010 & [-0.003, 0.028] & 0.0574 & 0.0923 & 0.0861 & n.s. & G \\
Small mol. & MoleculeNet & full-train common endpoints & AUC & TabPFN3 RDKit vs TabICL RDKit & endpoint & 667 & 9.84e-04 & 0.002 & [-0.003, 0.005] & 0.0288 & 0.0105 & 0.0575 & n.s. & G \\
Small mol. & MoleculeNet & full-train common endpoints & AUC & TabPFN3 RDKit vs ChemProp+CheMeleon & endpoint & 651 & 0.074 & 0.068 & [0.067, 0.081] & 1.10e-77 & 8.16e-106 & 6.59e-77 & *** & G \\
Small mol. & FS-Mol & full-train common endpoints & AUC & TabPFN3 ECFP+RDKit vs TabICL RDKit & endpoint & 157 & 0.009 & 0.007 & [-8.54e-04, 0.018] & 0.0224 & 0.0788 & 0.0575 & n.s. & G \\
Small mol. & FS-Mol & full-train common endpoints & AUC & TabPFN3 ECFP+RDKit vs ChemProp+CheMeleon & endpoint & 156 & 0.006 & 0.006 & [-0.004, 0.016] & 0.0947 & 0.0292 & 0.1136 & n.s. & G \\
Small mol. & DrugOOD & full-train OOD descriptive & OOD AUC & TabPFN3 ECFP+RDKit vs TabICL ECFP & shift & 3 & 0.013 & 0.017 & [-4.99e-04, 0.024] & 0.5000 & 1.0000 & 0.6667 & n.s. & H \\
Small mol. & DrugOOD & full-train ID/OOD gap descriptive & Gap & TabPFN3 ECFP+RDKit vs TabICL ECFP & shift & 3 & -0.054 & -0.056 & [-0.058, -0.050] & 0.2500 & 0.2500 & 0.5000 & n.s. & H \\
Small mol. & DrugOOD & full-train OOD descriptive & OOD AUC & TabPFN3 ECFP+RDKit vs ChemProp+CheMeleon & shift & 3 & -3.33e-04 & 0.002 & [-0.012, 0.009] & 1.0000 & 1.0000 & 1.0000 & n.s. & H \\
Small mol. & DrugOOD & full-train ID/OOD gap descriptive & Gap & TabPFN3 ECFP+RDKit vs ChemProp+CheMeleon & shift & 3 & -0.022 & -0.024 & [-0.034, -0.008] & 0.2500 & 0.2500 & 0.5000 & n.s. & H \\
Small mol. & DrugOOD & Figure 7; ECFP & AUC & TabPFN3 vs TabICL & endpoint & 3 & -0.031 & -0.024 & [-0.054, -0.014] & 0.2500 & 0.2500 & 0.4286 & n.s. & E \\
Small mol. & DrugOOD & Figure 7; ECFP+RDKit & AUC & TabPFN3 vs TabICL & endpoint & 3 & -0.017 & -0.016 & [-0.033, -0.002] & 0.2500 & 0.2500 & 0.4286 & n.s. & E \\
Small mol. & DrugOOD & Figure 7; RDKit & AUC & TabPFN3 vs TabICL & endpoint & 3 & 0.004 & 0.003 & [-1.73e-04, 0.009] & 0.5000 & 1.0000 & 0.6000 & n.s. & E \\
Small mol. & FS-Mol & Figure 7; ECFP & AUC & TabPFN3 vs TabICL & endpoint & 157 & -0.002 & -0.004 & [-0.005, 6.71e-04] & 0.0737 & 0.0376 & 0.2211 & n.s. & E \\
Small mol. & FS-Mol & Figure 7; ECFP+RDKit & AUC & TabPFN3 vs TabICL & endpoint & 157 & 0.005 & 0.003 & [5.40e-04, 0.009] & 0.0502 & 0.0376 & 0.2007 & n.s. & E \\
Small mol. & FS-Mol & Figure 7; RDKit & AUC & TabPFN3 vs TabICL & endpoint & 157 & 2.48e-05 & -9.77e-04 & [-0.003, 0.003] & 0.7471 & 0.5233 & 0.8150 & n.s. & E \\
Small mol. & MoleculeNet & Figure 7; ECFP & AUC & TabPFN3 vs TabICL & endpoint & 709 & 5.75e-04 & -0.004 & [-0.006, 0.007] & 0.0996 & 0.0068 & 0.2390 & n.s. & E \\
Small mol. & MoleculeNet & Figure 7; ECFP+RDKit & AUC & TabPFN3 vs TabICL & endpoint & 695 & -0.005 & -6.89e-04 & [-0.009, 4.44e-05] & 0.3105 & 0.7045 & 0.4657 & n.s. & E \\
Small mol. & MoleculeNet & Figure 7; RDKit & AUC & TabPFN3 vs TabICL & endpoint & 709 & -0.007 & -0.003 & [-0.010, -0.003] & 1.46e-06 & 2.28e-04 & 1.75e-05 & *** & E \\
Small mol. & TDC ADMET & Figure 7; ECFP & AUC & TabPFN3 vs TabICL & endpoint & 13 & -0.010 & -4.14e-04 & [-0.022, 0.001] & 0.3757 & 0.5811 & 0.5010 & n.s. & E \\
Small mol. & TDC ADMET & Figure 7; ECFP+RDKit & AUC & TabPFN3 vs TabICL & endpoint & 13 & 0.012 & 0.017 & [0.003, 0.020] & 0.0215 & 0.2668 & 0.1289 & n.s. & E \\
Small mol. & TDC ADMET & Figure 7; RDKit & AUC & TabPFN3 vs TabICL & endpoint & 13 & 9.76e-04 & -9.44e-04 & [-0.004, 0.006] & 1.0000 & 1.0000 & 1.0000 & n.s. & E \\
\bottomrule
\end{longtable}
\end{landscape}

\section{Additional ProteinGym Diagnostics}

Table~\ref{tab:supp-esmc-ablation} isolates the zero-shot-score component while holding the ESMC600M encoder, mutation-aware features, learner, and evaluation files fixed. The addition improves MSE consistently by rank-based testing, but its Spearman effect depends on the split. In particular, it helps slightly under random and modulo splitting but reduces contiguous-split Spearman. This ablation does not isolate encoder scale from the remaining mutation-aware feature engineering.

\begin{table}[H]
\centering
\small
\begin{tabular}{llrrrr}
\toprule
\textbf{Metric} & \textbf{Official scheme} & \textbf{n} & \textbf{Mean $\Delta$} & \textbf{95\% CI} & \textbf{BH q} \\
\midrule
Spearman & Overall & 217 & -0.006 & [-0.010, -0.002] & 0.040 \\
Spearman & Random & 217 & 0.003 & [0.001, 0.005] & $3.3\times10^{-7}$ \\
Spearman & Modulo & 217 & 0.003 & [-0.001, 0.007] & $2.5\times10^{-4}$ \\
Spearman & Contiguous & 217 & -0.024 & [-0.035, -0.014] & $2.1\times10^{-4}$ \\
MSE & Overall & 217 & 0.016 & [0.001, 0.038] & $3.4\times10^{-19}$ \\
MSE & Random & 217 & 0.008 & [0.004, 0.013] & $1.6\times10^{-20}$ \\
MSE & Modulo & 217 & 0.030 & [-0.003, 0.084] & $5.1\times10^{-17}$ \\
MSE & Contiguous & 217 & 0.011 & [-0.002, 0.022] & $1.2\times10^{-11}$ \\
\bottomrule
\end{tabular}
\caption{Fixed-encoder ablation of adding zero-shot scores to the mutation-aware ESMC600M feature set. Positive deltas favor the feature set with zero-shot scores; for MSE, positive means lower error. Confidence intervals bootstrap the mean, while q-values correct Wilcoxon tests across these eight rows.}
\label{tab:supp-esmc-ablation}
\end{table}

Only two of the 217 assays in the reproduced random 5-fold comparison required PCA rescue. Table~\ref{tab:supp-rescue-sensitivity} shows that excluding both changes neither model ordering nor the substantive means.

\begin{table}[H]
\centering
\small
\begin{tabular}{llrrr}
\toprule
\textbf{Model} & \textbf{Assays} & \textbf{n} & \textbf{Mean Spearman} & \textbf{Mean MSE} \\
\midrule
TabPFN3 & All & 217 & 0.7667 & 0.3505 \\
TabPFN3 & Excluding HIS7 and SPG1 Olson & 215 & 0.7653 & 0.3533 \\
TabICL & All & 217 & 0.7529 & 0.3758 \\
TabICL & Excluding HIS7 and SPG1 Olson & 215 & 0.7514 & 0.3788 \\
\bottomrule
\end{tabular}
\caption{Sensitivity of the random 5-fold ProteinGym aggregate to the two PCA-rescued assays.}
\label{tab:supp-rescue-sensitivity}
\end{table}

Figure~\ref{fig:supp-pg-official} visualizes the official ProteinGym split-scheme comparison that is summarized in the main text. The gap between random and harder modulo/contiguous splits is visible for both tabular in-context learners.

\begin{figure}[H]
\centering
\includegraphics[width=0.95\textwidth]{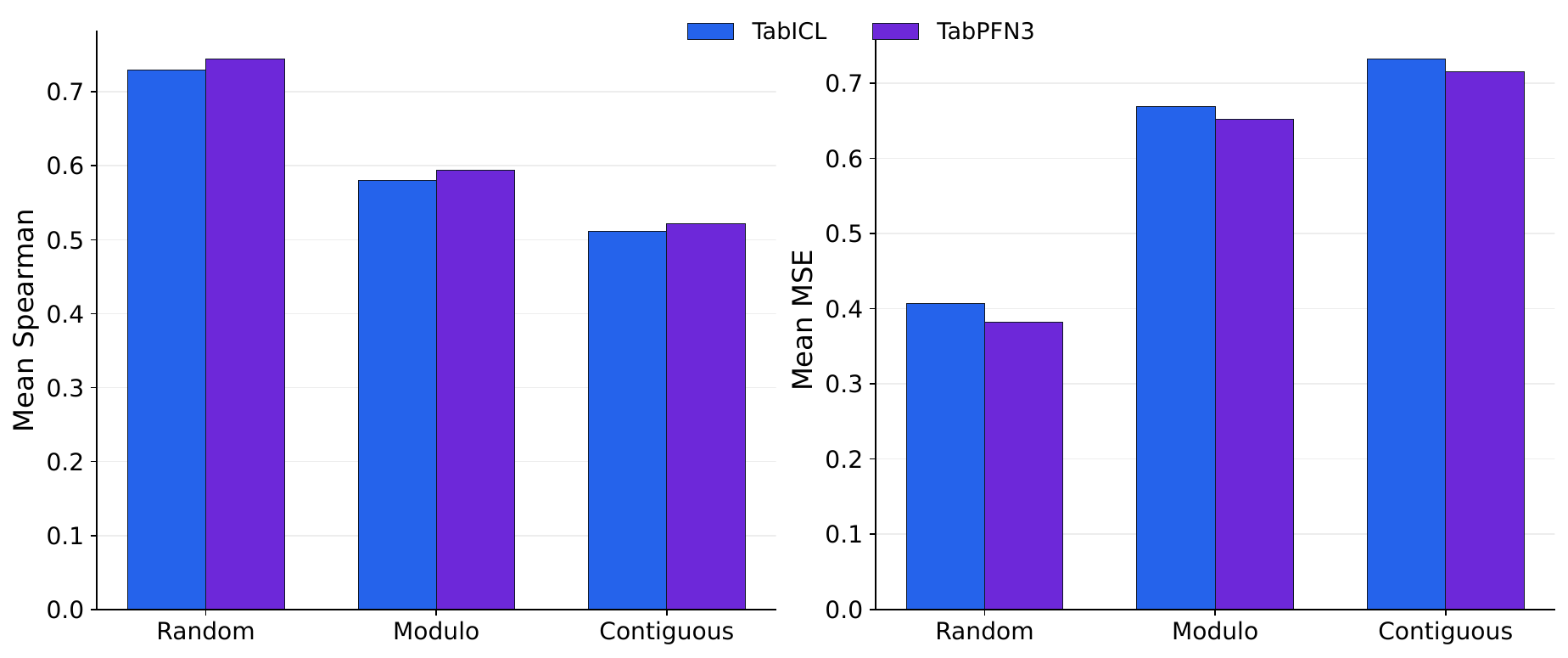}
\caption{ProteinGym official split-scheme performance. Random splits are easier than modulo and contiguous splits, emphasizing that random 5-fold numbers should not be treated as the only estimate of variant-effect generalization.}
\label{fig:supp-pg-official}
\end{figure}

Figure~\ref{fig:supp-pg-official-delta} shows paired fold-level differences between TabPFN3 and TabICL under official ProteinGym split schemes.

\begin{figure}[H]
\centering
\includegraphics[width=0.86\textwidth]{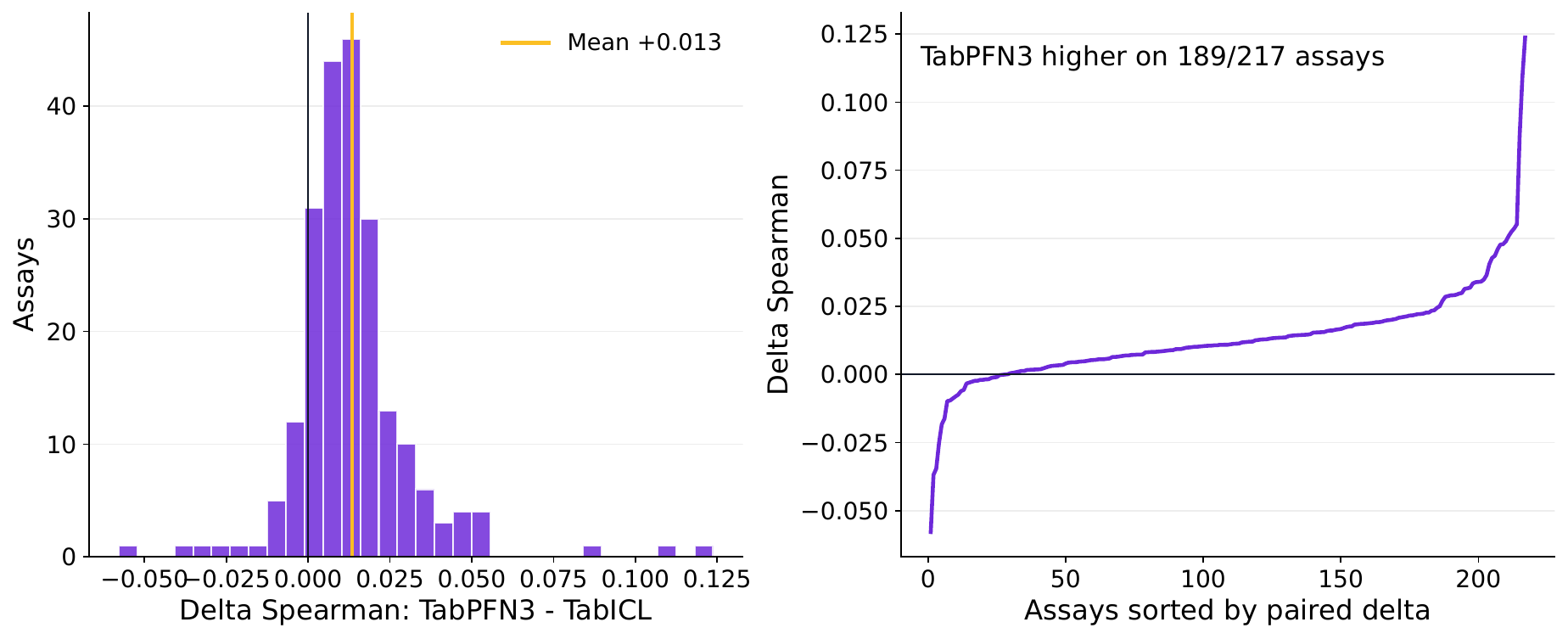}
\caption{Paired ProteinGym official-split differences. Positive values favor TabPFN3, negative values favor TabICL, after matching assay, split scheme, and fold.}
\label{fig:supp-pg-official-delta}
\end{figure}

Figure~\ref{fig:supp-pg-random-delta} gives the corresponding paired comparison for the random 5-fold full-data protocol.

\begin{figure}[H]
\centering
\includegraphics[width=0.86\textwidth]{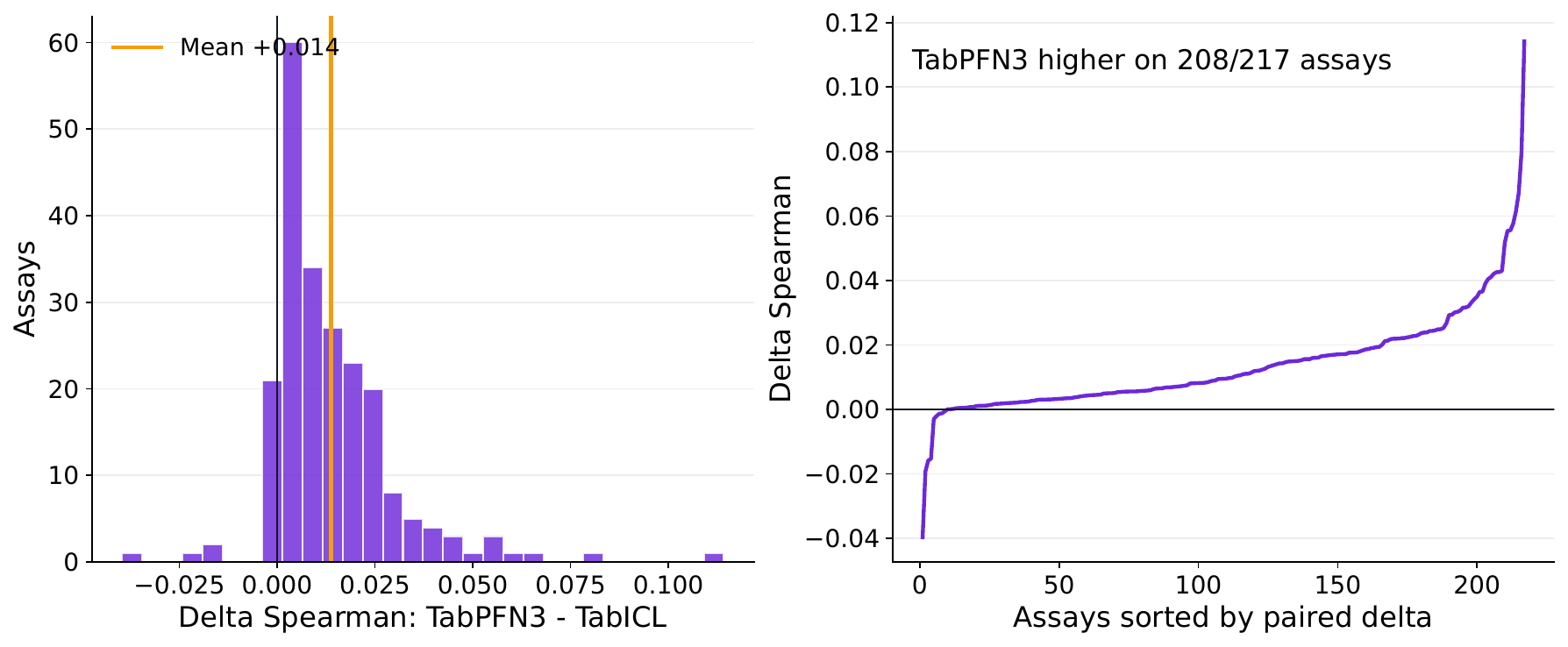}
\caption{Paired ProteinGym random 5-fold differences between TabPFN3 and TabICL. Positive values favor TabPFN3 and negative values favor TabICL.}
\label{fig:supp-pg-random-delta}
\end{figure}

Figure~\ref{fig:supp-pg-size} shows the assay-size diagnostic. This plot is exploratory: it checks whether performance differences are dominated by assay size rather than model or representation choice.

\begin{figure}[H]
\centering
\includegraphics[width=0.82\textwidth]{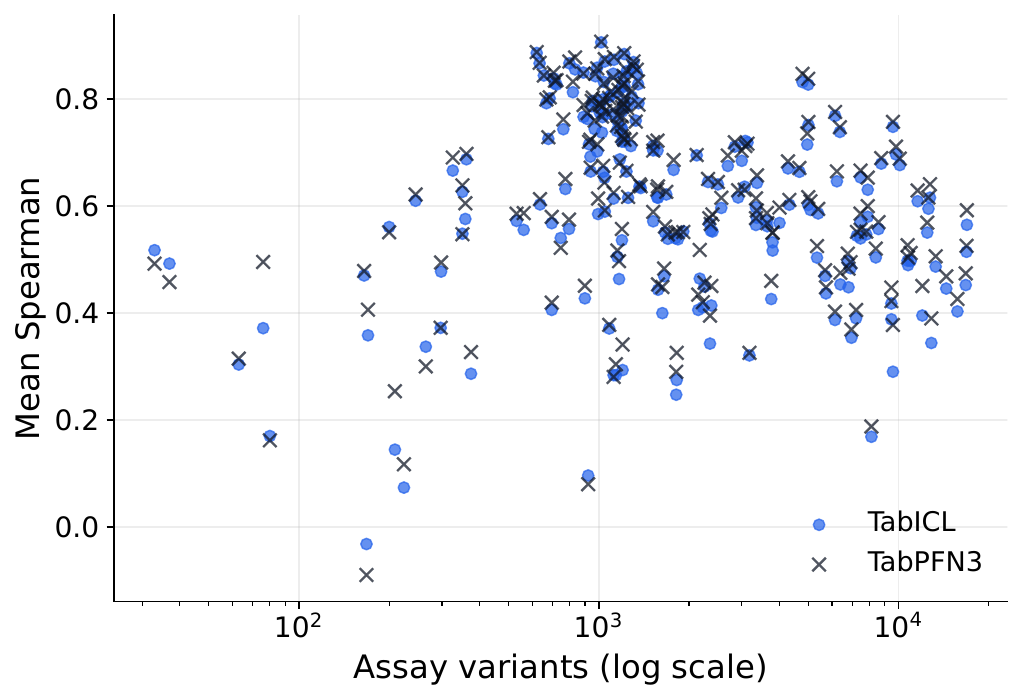}
\caption{ProteinGym assay-size diagnostic under the random 5-fold protocol. This is included as a diagnostic rather than a headline result.}
\label{fig:supp-pg-size}
\end{figure}

\section{PpEST TabFM Diagnostics}

The main manuscript includes TabFM only as a PpEST diagnostic. Table~\ref{tab:supp-tabfm-sweep} reports the full PCA sweep for TabFM on PpEST using ESMC300M sequence-level embeddings and the fixed full-training split. TabFM PCA128 gives the strongest TabFM result in this sweep, but it still does not exceed TabPFN3 in mean Spearman.

\begin{table}[H]
\centering
\small
\begin{tabular}{lrrrr}
\toprule
\textbf{Model / setting} & \textbf{Endpoints} & \textbf{Mean Spearman} & \textbf{Mean MSE} & \textbf{Seconds / endpoint} \\
\midrule
TabPFN3 & 5 & \textbf{0.660533} & 0.640193 & 18.932 \\
TabFM PCA16 & 5 & 0.632513 & 0.652681 & 45.653 \\
TabFM PCA32 & 5 & 0.647194 & 0.637768 & 76.278 \\
TabFM PCA64 & 5 & 0.644881 & 0.629834 & 92.652 \\
TabFM PCA128 & 5 & 0.649700 & \textbf{0.624843} & 105.442 \\
TabFM PCA500 & 5 & 0.641821 & 0.640622 & 360.129 \\
TabICL PCA500 & 5 & 0.643301 & 0.687250 & 21.460 \\
\bottomrule
\end{tabular}
\caption{PpEST full-training diagnostic sweep for TabFM using ESMC300M sequence-level embeddings. TabFM is competitive in some PCA settings but does not improve over TabPFN3 in mean Spearman.}
\label{tab:supp-tabfm-sweep}
\end{table}

Table~\ref{tab:supp-tabfm-per-endpoint-spearman} reports endpoint-level Spearman values for the TabFM PCA sweep.

\begin{table}[H]
\centering
\small
\begin{tabular}{lrrrrr}
\toprule
\textbf{Endpoint} & \textbf{PCA16} & \textbf{PCA32} & \textbf{PCA64} & \textbf{PCA128} & \textbf{PCA500} \\
\midrule
60c butyrate & 0.699526 & 0.715356 & 0.707840 & 0.706072 & 0.710907 \\
90c butyrate & 0.688162 & 0.722325 & 0.722423 & 0.711896 & 0.717843 \\
acetate & 0.595913 & 0.642562 & 0.629296 & 0.641771 & 0.646082 \\
butyrate & 0.541929 & 0.524301 & 0.528012 & 0.548861 & 0.507362 \\
octanoate & 0.637035 & 0.631428 & 0.636832 & 0.639902 & 0.626911 \\
\bottomrule
\end{tabular}
\caption{Endpoint-level Spearman for the TabFM PpEST PCA sweep.}
\label{tab:supp-tabfm-per-endpoint-spearman}
\end{table}

Table~\ref{tab:supp-tabfm-per-endpoint-mse} reports the corresponding endpoint-level MSE values.

\begin{table}[H]
\centering
\small
\begin{tabular}{lrrrrr}
\toprule
\textbf{Endpoint} & \textbf{PCA16} & \textbf{PCA32} & \textbf{PCA64} & \textbf{PCA128} & \textbf{PCA500} \\
\midrule
60c butyrate & 0.647077 & 0.561668 & 0.581760 & 0.573383 & 0.563566 \\
90c butyrate & 0.590467 & 0.582136 & 0.586242 & 0.620080 & 0.626269 \\
acetate & 0.713357 & 0.695283 & 0.647309 & 0.657457 & 0.675629 \\
butyrate & 0.530987 & 0.567352 & 0.581027 & 0.541406 & 0.595882 \\
octanoate & 0.781518 & 0.782402 & 0.752834 & 0.731889 & 0.741761 \\
\bottomrule
\end{tabular}
\caption{Endpoint-level MSE for the TabFM PpEST PCA sweep.}
\label{tab:supp-tabfm-per-endpoint-mse}
\end{table}

No full-train PpEST comparison is statistically significant after correction at five endpoints. Table~\ref{tab:supp-ppest-paired} reports the principal descriptive paired effects. A two-sided exact sign or Wilcoxon test cannot reach $p<0.05$ with only five non-zero pairs, even if one method wins all five endpoints.

\begin{table}[H]
\centering
\scriptsize
\begin{tabular}{llrrrr}
\toprule
\textbf{Metric} & \textbf{Comparison} & \textbf{n} & \textbf{Mean $\Delta$} & \textbf{95\% CI} & \textbf{BH q} \\
\midrule
Spearman & TabPFN3 ESMC300M vs TabICL no-PCA & 5 & 0.013 & [0.003, 0.023] & 0.250 \\
MSE & TabPFN3 ESMC300M vs TabICL no-PCA & 5 & -0.003 & [-0.032, 0.029] & 1.000 \\
Spearman & TabPFN3 ESMC300M vs fine-tuned ESM & 5 & 0.072 & [0.044, 0.113] & 0.156 \\
MSE & TabPFN3 ESMC300M vs fine-tuned ESM & 5 & 0.050 & [0.019, 0.091] & 0.156 \\
Spearman & TabPFN3 ESMC300M vs Sequence CNN & 5 & 0.139 & [0.116, 0.166] & 0.156 \\
MSE & TabPFN3 ESMC300M vs Sequence CNN & 5 & 0.060 & [-0.022, 0.155] & 0.521 \\
Spearman & TabPFN3 ESMC300M vs ESMC600M & 5 & 0.002 & [-0.004, 0.007] & 0.833 \\
MSE & TabPFN3 ESMC300M vs ESMC600M & 5 & -0.009 & [-0.020, 0.005] & 0.536 \\
Spearman & TabICL no-PCA vs PCA500 & 5 & 0.004 & [-0.003, 0.013] & 1.000 \\
MSE & TabICL no-PCA vs PCA500 & 5 & 0.050 & [-0.034, 0.170] & 1.000 \\
\bottomrule
\end{tabular}
\caption{Exploratory endpoint-level PpEST paired comparisons. Positive deltas favor the first method, including lower MSE.}
\label{tab:supp-ppest-paired}
\end{table}

\section{Limited TabFM ProteinGym Feasibility Diagnostic}

We also ran a limited TabFM PCA32 feasibility diagnostic on ProteinGym. This run was not completed as a benchmark result and should not be interpreted as leaderboard-comparable performance. It is included here only to document why TabFM was not pursued as a full ProteinGym method in the main manuscript.

\begin{table}[H]
\centering
\scriptsize
\resizebox{\textwidth}{!}{%
\begin{tabular}{rrrrrrrrr}
\toprule
\textbf{Folds} & \textbf{TabFM Spearman} & \textbf{TabPFN3 Spearman} & \textbf{$\Delta$ Spearman} & \textbf{Wins} & \textbf{Losses} & \textbf{Tie/undef.} & \textbf{TabFM MSE} & \textbf{TabPFN3 MSE} \\
\midrule
229 & 0.4596 & 0.5091 & -0.0495 & 88 & 138 & 3 & 1.1378 & 0.9626 \\
\bottomrule
\end{tabular}
}
\caption{Partial TabFM PCA32 ProteinGym feasibility diagnostic. Values compare completed TabFM folds with matched TabPFN3 baseline folds. Tie/undef. counts folds with tied or undefined paired Spearman comparisons. The run was incomplete and included operational failures, so it is reported only as exploratory evidence.}
\label{tab:supp-tabfm-pg-overall}
\end{table}

\begin{table}[H]
\centering
\scriptsize
\resizebox{\textwidth}{!}{%
\begin{tabular}{lrrrrrrr}
\toprule
\textbf{Split} & \textbf{Folds} & \textbf{TabFM Spearman} & \textbf{TabPFN3 Spearman} & \textbf{$\Delta$ Spearman} & \textbf{Wins} & \textbf{Losses} & \textbf{Tie/undef.} \\
\midrule
Random & 83 & 0.6351 & 0.6909 & -0.0557 & 21 & 60 & 2 \\
Modulo & 79 & 0.3848 & 0.4473 & -0.0625 & 37 & 42 & 0 \\
Contiguous & 67 & 0.3302 & 0.3567 & -0.0265 & 30 & 36 & 1 \\
\bottomrule
\end{tabular}
}
\caption{Split-level view of the partial TabFM PCA32 ProteinGym feasibility diagnostic. Tie/undef. counts folds with tied or undefined paired Spearman comparisons.}
\label{tab:supp-tabfm-pg-splits}
\end{table}

The partial ProteinGym diagnostic does not suggest a compelling TabFM advantage. Because it is incomplete, it is weaker evidence than the complete PpEST TabFM sweep. Its main value is as a model-selection guardrail: larger or newer tabular foundation models do not automatically improve biomolecular rank prediction under the tested representations and resource constraints.

\section{Additional Molecular Diagnostics}

Figure~\ref{fig:supp-mol-fulltrain-summary} gives the full-training small-molecule overview corresponding to the endpoint summary in the main manuscript. Figure~\ref{fig:supp-mol-fewshot-coverage} shows few-shot molecular coverage across the benchmark families, and Figure~\ref{fig:supp-mol-fulltrain-coverage} shows full-training job/file coverage. These diagnostics are useful for interpreting missing or incomplete endpoint families, especially resource-limited MoleculeNet PCBA full-training runs. The coverage counts in Figure~\ref{fig:supp-mol-fulltrain-coverage} count completed output files, whereas the main manuscript's MoleculeNet coverage row reports endpoint rows with defined ROC-AUC after metric filtering. For example, TabPFN3 RDKit has 678 completed MoleculeNet outputs in the coverage diagnostic but 667 defined ROC-AUC rows in the main summary.

\begin{figure}[H]
\centering
\includegraphics[width=0.98\textwidth]{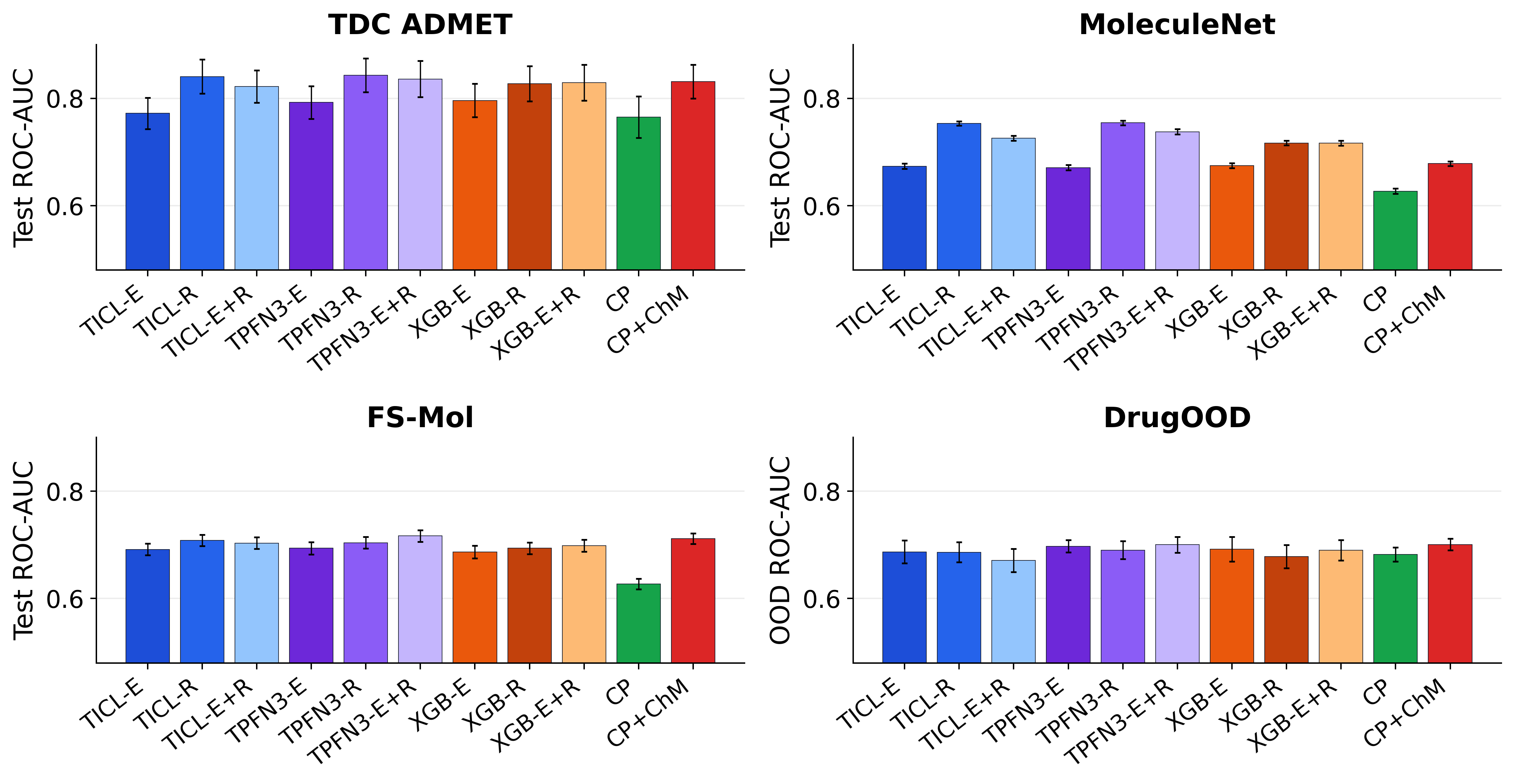}
\caption{Full-training small-molecule benchmark summary. Bars show mean ROC-AUC across tasks; DrugOOD uses OOD-test ROC-AUC and the other benchmark families use benchmark test ROC-AUC.}
\label{fig:supp-mol-fulltrain-summary}
\end{figure}

\begin{table}[H]
\centering
\small
\begin{tabular}{lrrrr}
\toprule
\textbf{MoleculeNet full-train pair} & \textbf{Outputs} & \textbf{Defined} & \textbf{Undefined} & \textbf{Missing of 806} \\
\midrule
TabICL ECFP & 713 & 701 & 12 & 93 \\
TabICL RDKit & 678 & 667 & 11 & 128 \\
TabICL ECFP+RDKit & 678 & 667 & 11 & 128 \\
TabPFN3 ECFP & 678 & 667 & 11 & 128 \\
TabPFN3 RDKit & 678 & 667 & 11 & 128 \\
TabPFN3 ECFP+RDKit & 678 & 667 & 11 & 128 \\
\bottomrule
\end{tabular}
\caption{MoleculeNet full-train ROC-AUC accounting for the tabular predictor-representation pairs. Undefined denotes a completed endpoint whose test split has one class. Missing TabPFN3 endpoints are the 128 resource-infeasible PCBA outputs and are not scored as failures.}
\label{tab:supp-mol-undefined}
\end{table}

On the exact common set of 667 defined RDKit endpoints, TabPFN3 has mean ROC-AUC 0.7544 and TabICL has 0.7534, a mean paired difference of 0.0010 with bootstrap 95\% CI $[-0.0027,0.0046]$ and corrected $q=0.0575$. We therefore do not claim a clear MoleculeNet winner. Across all model families in the stored result registry, undefined ROC-AUC units occur only for MoleculeNet: 2,180 of 202,126 few-shot evaluation units and 404 of 28,611 full-train units after duplicate exports are collapsed. TDC ADMET, FS-Mol, and DrugOOD contain no undefined units in the corresponding stored rows.

\begin{figure}[H]
\centering
\includegraphics[width=0.9\textwidth]{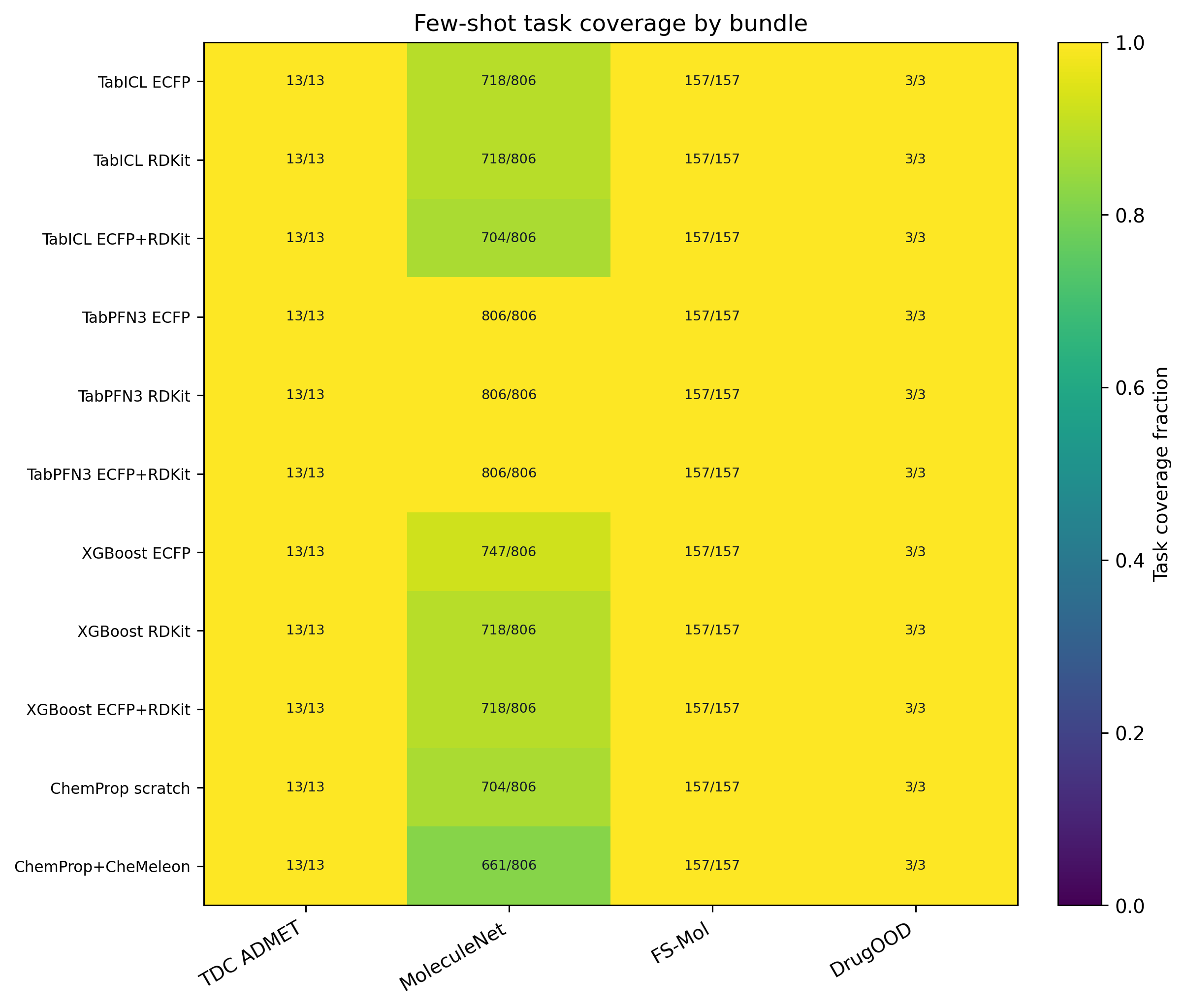}
\caption{Few-shot molecular benchmark coverage diagnostic.}
\label{fig:supp-mol-fewshot-coverage}
\end{figure}

\begin{figure}[H]
\centering
\includegraphics[width=0.9\textwidth]{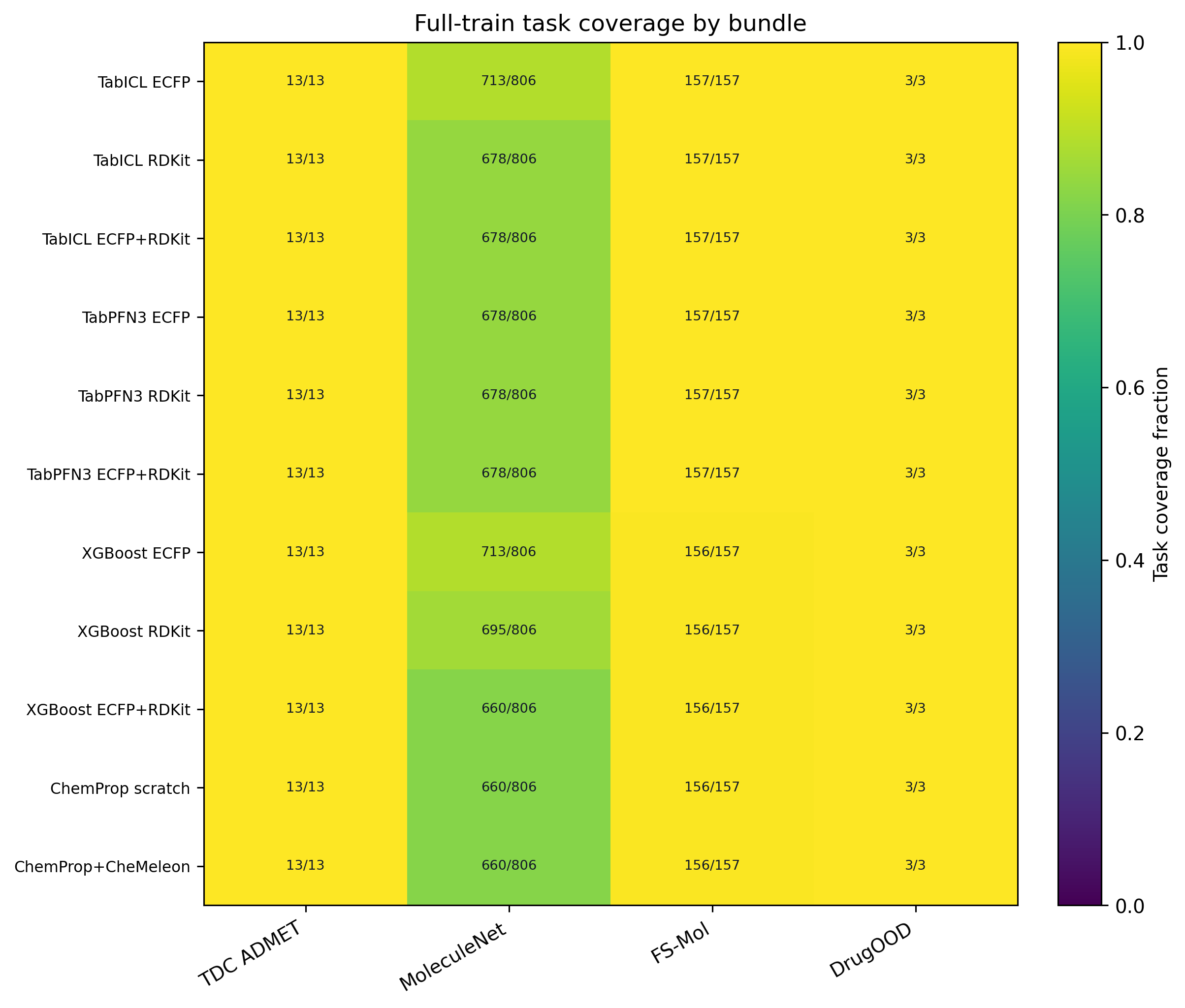}
\caption{Full-training molecular benchmark coverage diagnostic. Counts denote completed output files before filtering undefined ROC-AUC rows; main-text coverage counts report defined ROC-AUC rows after metric filtering.}
\label{fig:supp-mol-fulltrain-coverage}
\end{figure}

\begin{figure}[H]
\centering
\includegraphics[width=0.82\textwidth]{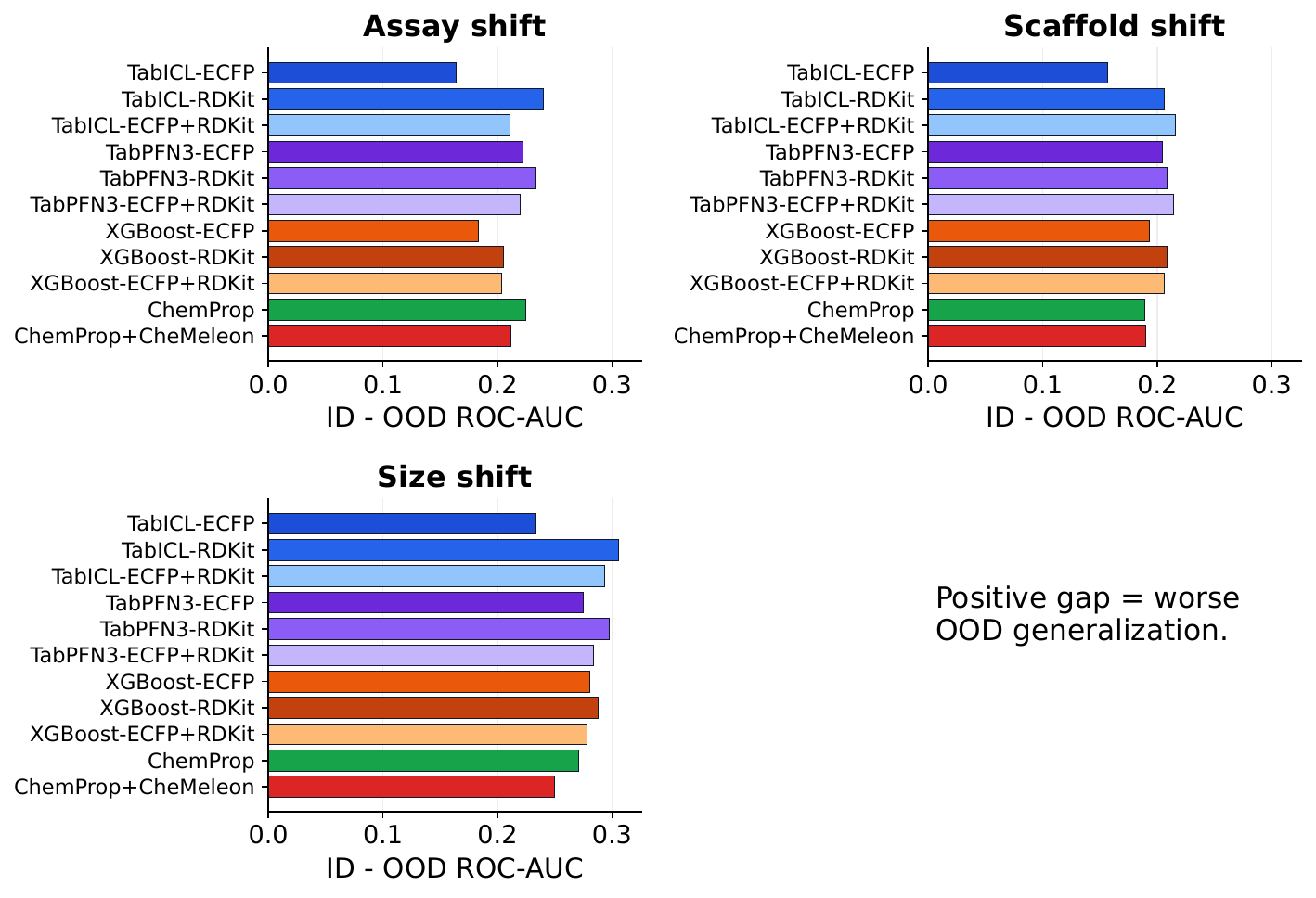}
\caption{DrugOOD full-train ID/OOD generalization gap. Bars report ID-test ROC-AUC minus OOD-test ROC-AUC for each predictor--representation pair on assay, scaffold, and molecular-size shifts. Larger positive values indicate a larger OOD performance drop. With only three shifts, this result is descriptive.}
\label{fig:supp-drugood-gap}
\end{figure}

Figure~\ref{fig:supp-mol-reference-context} provides additional external molecular benchmark context. It is included as orientation rather than as a direct like-for-like comparison, because benchmark construction, representation, and training protocols vary substantially across sources. TDC top and reference values are taken from Therapeutics Data Commons ADMET benchmark pages (\url{https://tdcommons.ai/benchmark/admet_group/overview/}, accessed July 2026). MoleculeNet reference values are from the MoleculeNet benchmark paper and associated benchmark summary tables, and DrugOOD values are from the DrugOOD LBAP-core IC50 reference tables. The source tables used to generate this figure are included as ancillary files with the arXiv submission under \texttt{source\_tables/}.

\begin{figure}[H]
\centering
\includegraphics[width=0.9\textwidth]{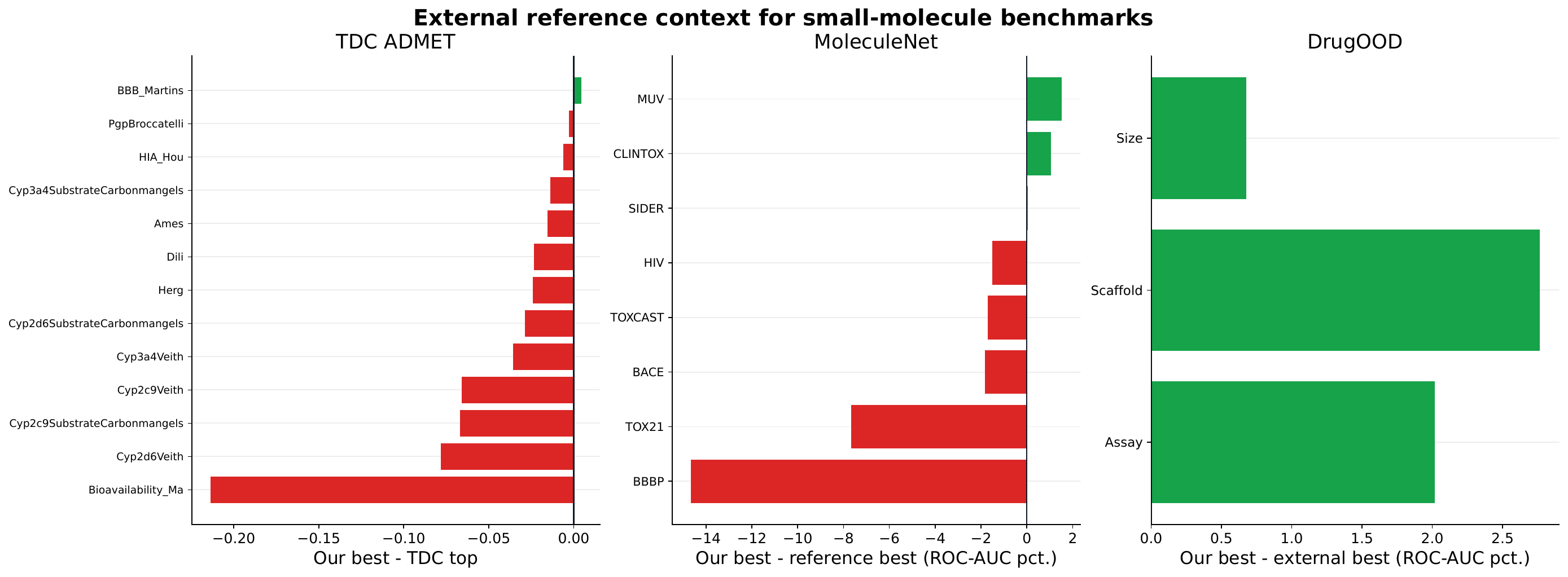}
\caption{External molecular benchmark context for selected settings. TDC values use the public TDC ADMET benchmark pages accessed July 2026; MoleculeNet and DrugOOD values use the corresponding benchmark/reference tables included in \texttt{source\_tables/}. These values should be read as orientation rather than as a fully controlled comparison.}
\label{fig:supp-mol-reference-context}
\end{figure}

\section{Reviewer-Triggered Extra Experiments}

This section collects the additional analyses run specifically in response to reviewer concerns. They are included to document the evidence behind the rebuttal, but they should be read as revision support rather than as new headline benchmarks.

\subsection{Label-Agnostic ProteinGym Few-Shot Support Sweep}

Table~\ref{tab:supp-random-support} reports the random-support ProteinGym control. Unlike the submitted quantile-bin support protocol, this sweep samples support points without using labels. The strongest $k=64$ conclusion remains, while $k=8$ Spearman is mixed and therefore should be qualified.

\begin{table}[H]
\centering
\footnotesize
\setlength{\tabcolsep}{5pt}
\begin{tabular}{llrrrrrr}
\toprule
\textbf{k} & \textbf{Metric} & \textbf{n} & \textbf{Mean $\Delta$} & \textbf{95\% CI} & \textbf{BH q} & \textbf{TabPFN3 wins} & \textbf{TabICL wins} \\
\midrule
8 & Spearman & 217 & -0.0047 & [-0.0095, 0.0001] & 0.0086 & 87 & 130 \\
8 & MSE & 217 & 0.0754 & [0.0414, 0.1175] & $2.0\times10^{-10}$ & 150 & 67 \\
64 & Spearman & 216 & 0.0246 & [0.0213, 0.0281] & $1.8\times10^{-28}$ & 191 & 25 \\
64 & MSE & 216 & 0.0429 & [0.0377, 0.0481] & $2.0\times10^{-32}$ & 201 & 15 \\
\bottomrule
\end{tabular}
\caption{ProteinGym random-support control comparing TabPFN3 and TabICL. Values aggregate five random support repeats per assay before paired testing. Positive deltas favor TabPFN3; for MSE, positive means lower MSE for TabPFN3.}
\label{tab:supp-random-support}
\end{table}

\subsection{TabPFN3 ESMC600M PCA500 Control}

Table~\ref{tab:supp-tabpfn3-pca500-control} tests whether compressing ESMC600M features to PCA500 is neutral for TabPFN3. It is not: full mutation-aware ESMC600M features perform better than the PCA500 version, supporting our statement that TabICL-PCA500 is a practical feasibility setting rather than a controlled dimension-matched ablation.

\begin{table}[H]
\centering
\small
\begin{tabular}{llrrrrr}
\toprule
\textbf{Metric} & \textbf{Setting} & \textbf{n} & \textbf{Full 9303d} & \textbf{PCA500} & \textbf{Mean $\Delta$} & \textbf{BH q} \\
\midrule
Spearman & Overall & 217 & 0.662 & 0.628 & 0.0343 & $1.1\times10^{-23}$ \\
MSE & Overall & 217 & 0.537 & 0.601 & 0.0638 & $8.0\times10^{-25}$ \\
Spearman & Random & 217 & 0.771 & 0.763 & 0.0075 & $1.4\times10^{-8}$ \\
MSE & Random & 217 & 0.368 & 0.381 & 0.0131 & $9.7\times10^{-9}$ \\
Spearman & Modulo & 217 & 0.605 & 0.557 & 0.0475 & $1.2\times10^{-22}$ \\
MSE & Modulo & 217 & 0.675 & 0.759 & 0.0840 & $1.1\times10^{-25}$ \\
Spearman & Contiguous & 217 & 0.550 & 0.502 & 0.0477 & $7.2\times10^{-23}$ \\
MSE & Contiguous & 217 & 0.729 & 0.824 & 0.0944 & $1.1\times10^{-23}$ \\
\bottomrule
\end{tabular}
\caption{Matched TabPFN3 ESMC600M full-dimensional versus PCA500 control across official ProteinGym split schemes. Positive deltas favor full-dimensional ESMC600M; for MSE, positive means lower MSE for the full-dimensional features.}
\label{tab:supp-tabpfn3-pca500-control}
\end{table}

\subsection{Stronger Matched ProteinGym Baseline Controls}

Table~\ref{tab:supp-stronger-baselines} adds stronger matched ProteinGym controls on the same 217-assay random 5-fold ESMC300M protocol. These replace the role of the untuned Ridge and RBF diagnostic rows when discussing competitive supervised baselines.

\begin{table}[H]
\centering
\small
\begin{tabular}{lrr}
\toprule
\textbf{Model} & \textbf{Mean Spearman} & \textbf{Mean MSE} \\
\midrule
TabPFN3 & 0.767 & 0.351 \\
Fixed-encoder ESMC300M MLP & 0.722 & 0.405 \\
Training-fold-tuned Ridge & 0.712 & 0.424 \\
\bottomrule
\end{tabular}
\caption{Stronger ProteinGym random 5-fold baseline controls using the same ESMC300M assay-level protocol. TabPFN3 remains higher than the fixed-encoder MLP on the same assays, with paired mean $\Delta$Spearman $=0.0448$ and paired mean $\Delta$MSE $=0.0546$ after correction.}
\label{tab:supp-stronger-baselines}
\end{table}

\subsection{PpEST Identity-Clustered Split}

Table~\ref{tab:supp-ppest-clustered} reports an exploratory PpEST identity-clustered robustness check. Each CD-HIT sequence-identity cluster is confined to one fold. Because PpEST has only five endpoints, these results are descriptive and should not be interpreted as definitive statistical evidence.

\begin{table}[H]
\centering
\small
\begin{tabular}{lrrrr}
\toprule
\textbf{Model} & \textbf{Endpoints} & \textbf{Folds} & \textbf{Mean Spearman} & \textbf{Mean MSE} \\
\midrule
TabPFN3 & 5 & 25 & 0.543 & 0.879 \\
TabICL & 5 & 25 & 0.532 & 0.886 \\
Fixed-encoder ESMC300M MLP & 5 & 25 & 0.435 & 0.968 \\
Training-fold-tuned Ridge & 5 & 25 & 0.428 & 0.956 \\
HistGradientBoosting & 5 & 25 & 0.369 & 1.014 \\
\bottomrule
\end{tabular}
\caption{Exploratory PpEST identity-clustered split. Absolute performance decreases relative to the random split, but TabPFN3 and TabICL remain strongest among the tested predictors.}
\label{tab:supp-ppest-clustered}
\end{table}

\section{Interpretation Of Diagnostic Results}

The appendix diagnostics support three points.

First, ProteinGym performance depends strongly on split scheme. Random folds are substantially easier than modulo or contiguous folds, so official holdout schemes are important for interpreting variant-effect generalization.

Second, representation changes are task dependent. Mutation-aware ESMC600M features improve TabPFN3 on ProteinGym, but sequence-level ESMC600M embeddings do not improve PpEST Spearman over ESMC300M.

Third, TabFM provides a useful negative diagnostic. On PpEST, it is competitive in some PCA settings but does not exceed TabPFN3 in mean Spearman. On the limited ProteinGym feasibility run, it did not show a compelling enough advantage to justify a full benchmark-scale run. This supports the main manuscript's claim that performance depends on the learner-representation pairing rather than model scale alone.
\endgroup

\end{document}